%% file: main.tex
\documentclass[letterpaper]{article}
\usepackage[preprint]{aaai2027}
\usepackage[hyphens]{url}
\usepackage{graphicx}
\usepackage{natbib}
\usepackage{caption}
\usepackage{enumitem}
\usepackage{amsfonts}
\usepackage{amsmath}
\usepackage{amssymb}
\usepackage{booktabs}
\usepackage{multirow}
\usepackage{tabularx}
\usepackage{makecell}
\usepackage{array}
\usepackage[table]{xcolor}
\usepackage{pifont}
\usepackage[ruled,vlined]{algorithm2e}
\DontPrintSemicolon

\definecolor{oursblue}{RGB}{232,232,255}
\newcommand{\cmark}{\textcolor{red}{\ding{51}}}
\newcommand{\xmark}{\textcolor{green}{\ding{55}}}

\title{Beyond Static Anchors: Bounded Prototype Conditioning for Language-Free Medical Anomaly Detection}

\author{
    Yibo Wan\textsuperscript{\rm 1},
    Jinyu Cai\textsuperscript{\rm 2}\corresponding,
    See-Kiong Ng\textsuperscript{\rm 2}
}
\affiliations{
    \textsuperscript{\rm 1}School of Computing, National University of Singapore, Singapore\\
    \textsuperscript{\rm 2}Institute of Data Science, National University of Singapore, Singapore\\
    yibo.wan19@outlook.com, \{jinyucai,seekiong\}@nus.edu.sg
}

\begin{document}

\maketitle

\begin{abstract}
Medical anomaly detection identifies abnormal images and localizes lesions under scarce supervision while generalizing across organs and modalities. Existing CLIP-based methods reduce annotation requirements through vision--language alignment, but their normal and abnormal references, whether text prompts or learned visual tokens, remain fixed across test images. Such static references may not transfer reliably to unseen targets in a cross-domain medical imaging scenario. To address this, we propose ReCAP, a language-free framework that replaces static anchors with input-conditioned visual prototypes. ReCAP re-centers separated normal and abnormal prototypes for each image through a bounded gated modulation, enabling query-adaptive anomaly scoring while constraining context-induced prototype drift. For the few-shot setting, we introduce a non-parametric normal-reference memory to preserve instance-level target-domain variation and complement the conditional prototype branch. Across six medical benchmarks, ReCAP achieves the best image-level AUROC on all zero-shot and 23 of 24 few-shot settings, and the best zero-shot pixel-level AUROC on all three segmentation datasets. Particularly, it reduces inference latency by over 70\% compared to the fastest baseline, without text prompts or test-time gradient updates.
\end{abstract}

\section{Introduction}
\label{sec:intro}
Medical anomaly detection (AD) aims to identify abnormal images and localize lesion regions under limited supervision~\cite{cai2025medianomaly,woll22Diffusion}. This problem is especially important in medical imaging, where lesions are rare, diverse, and expensive to annotate. Therefore, a practical medical detector must operate under scarce supervision while remaining transferable across organs and modalities such as brain MRI, liver CT, and chest X-ray. This cross-domain setting is particularly challenging: normal appearance changes substantially across anatomical structures and imaging protocols, so an anomaly reference learned from source domains may be poorly calibrated for an unseen target domain.

Recent CLIP-based anomaly detectors reduce annotation cost by exploiting transferable vision--language representations~\cite{radford2021learning}. Prompt-based methods define normal and abnormal semantics with hand-crafted or learnable text prompts, while visual adaptation methods use multi-level CLIP image features for anomaly classification and localization~\cite{huang2024adapting,shiri2025madclip,ma2025aa}. More recent language-free methods, such as VisualAD~\cite{hou2026visualad} and UniADet~\cite{gao2026one}, remove the text encoder and learn normal/abnormal visual tokens directly in the image feature space. Although these methods differ in how they construct anomaly anchors, most of them share a common assumption: the normal/abnormal reference is fixed after training and reused for all test images. Such a static scoring reference can be brittle in cross-domain medical AD, where the feature-space location of normality may vary sharply across target domains.

This static-anchor formulation leads to three limitations. First, text prompts such as ``normal'' and ``abnormal'' provide coarse semantic supervision and are sensitive to wording, whereas medical lesions are often subtle, localized, and difficult to describe with generic language. Second, removing text does not by itself solve cross-domain miscalibration, because learned visual anchors can still define a fixed anomaly boundary that is misplaced for an unseen target. Third, compact global anchors cannot preserve the diverse instance-level normal patterns needed for fine-grained localization, especially when only a few target-domain support images are available. These limitations suggest that robust medical AD requires not only language-free anomaly anchors, but also a mechanism that adapts the anomaly boundary to each input while preserving normal/abnormal separability.

\begin{figure*}[t]
    \centering
    \includegraphics[width=0.85\textwidth]{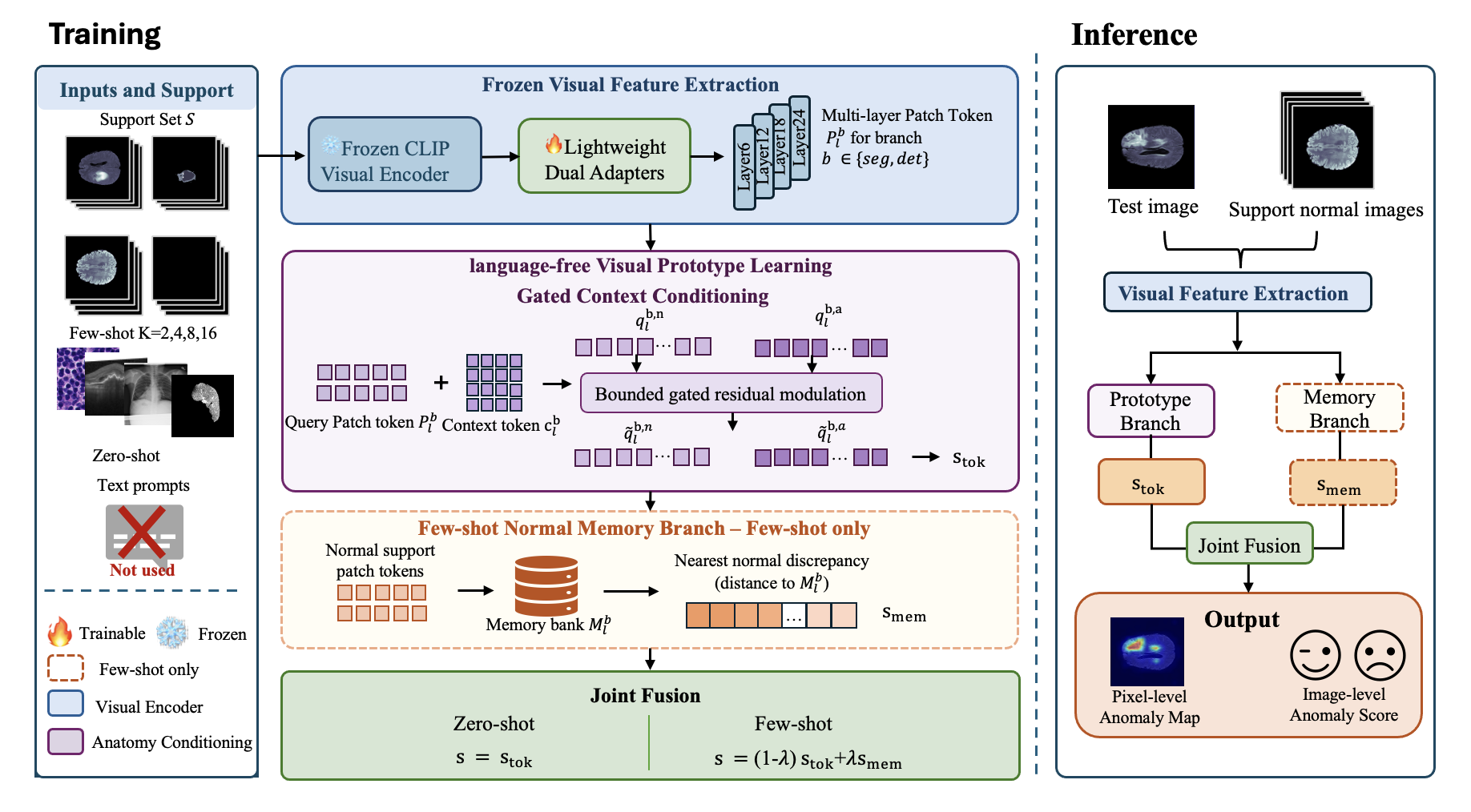}
 \caption{Overview of ReCAP. Instead of aligning visual features with hand-crafted text prompts, our framework learns normal and abnormal visual prototypes directly in the CLIP visual space. The prototypes are further modulated by query-specific anatomical context to produce adaptive visual anchors for anomaly scoring. In the few-shot setting, an auxiliary normal memory bank provides instance-level normal references, while the zero-shot setting relies solely on the learned visual prototypes.}
    \label{fig:framework}
\end{figure*}

To address these challenges, we propose \textbf{Re-C}entered \textbf{A}nomaly \textbf{P}rototypes (\textbf{ReCAP}), a language-free framework for zero- and few-shot medical anomaly detection, as illustrated in Figure~\ref{fig:framework}. The central idea is to replace a static anomaly reference with an input-conditioned visual prototype boundary. ReCAP re-centers a learned and separated prototype pair for each image according to its global visual context, producing an image-specific boundary without text prompts or test-time gradient updates. This re-centering is deliberately bounded to mitigate excessive prototype drift, because the conditioning signal is pooled from the input image itself and may be contaminated by the very lesion we aim to detect. Therefore, the boundary could shift toward anomalous content and explain it away as normal if left unconstrained. Together with a boundary-separation regularizer, the bounded design allows the prototype pair to adapt toward the current domain while preserving normal/abnormal discrimination, providing a controlled, gradient-free adaptation mechanism that supports zero-shot transfer. For few-shot detection, ReCAP further uses target-domain support images to initialize target-specific prototypes and builds a non-parametric normal-reference memory from normal support samples, preserving instance-level normal variation that complements the conditional prototype branch. Multi-layer fusion combines evidence from different CLIP layers and integrates prototype-based discrimination with memory-based deviation, producing both image-level anomaly scores and pixel-level anomaly maps. Built on a frozen CLIP visual encoder with lightweight adapters, ReCAP remains language-free, adaptive, and efficient. Our main contributions are summarized as follows:
\begin{itemize}[leftmargin=*]
\item We identify static anomaly anchors as a key limitation of CLIP-based medical AD under cross-domain shifts, and propose input-conditioned visual prototypes for language-free image-specific boundary prediction.
\item We introduce bounded gated prototype modulation to re-centers normal/abnormal visual prototypes according to input context while mitigating uncontrolled prototype drift.
\item We incorporate normal-reference memory and multi-layer fusion for few-shot detection and localization, achieving state-of-the-art image- and pixel-level performance with improved inference efficiency across six benchmarks.
\end{itemize}

\section{Related Work}
\label{sec:formatting}
\paragraph{Traditional Medical Anomaly Detection.}
Medical anomaly detection has been studied through reconstruction-based, feature embedding-based, and supervised paradigms~\cite{deng2022anomaly,he2024diffusion,sabokrou2018adversarially,wang2025distribution}. Reconstruction methods learn normal-image distributions and detect anomalies by reconstruction discrepancy~\cite{schlegl2017unsupervisedanomalydetectiongenerative,zong2018deep,gong2019memorizing,gudovskiy2022cflow,li2025one}, while embedding-based methods model normality in a representation space using one-class objectives, memory banks, or distributional estimates~\cite{roth2022towards,liu2023simplenet,koshil2024anomalouspatchcore,wang2025distribution}. Supervised methods can provide stronger discriminative signals but require costly anomaly annotations and often generalize poorly to unseen anomaly types~\cite{wang2025distribution,rolih2024supersimplenet}. These limitations are especially pronounced in medical imaging, where abnormal samples are scarce and normal anatomy varies substantially across organs and modalities.

\paragraph{CLIP-based Anomaly Detection.}
CLIP-based anomaly detectors reduce annotation cost by transferring vision--language representations to zero-/few-shot anomaly detection~\cite{radford2021learning,huang2024adapting,zhang2024mediclip}. Most methods define anomaly semantics through textual prompts or learnable text-side representations, and score images or patches by contrasting normal and abnormal prompt similarities~\cite{wang2022medclip,zhou2025ultraad,ma2025aa,cao2024adaclip,shiri2025madclip,huang2025iqe,zhang2024mediclip}. In medical imaging, this paradigm has been extended with medical vision--language pretraining, multi-level visual adaptation, and stronger normal/abnormal semantic separation~\cite{huang2024adapting,ma2025aa,zhou2023anomalyclip,shiri2025madclip}. However, prompt-based methods remain sensitive to textual descriptions, whereas many medical anomalies are subtle structural deviations that are difficult to capture with generic words such as \emph{normal} and \emph{abnormal}.

\paragraph{Language-free CLIP-based Anomaly Detection.}
Recent language-free methods remove the dependence on textual prompts by learning anomaly semantics directly in the CLIP visual space~\cite{hou2026visualad,gao2026one}. VisualAD~\cite{hou2026visualad} replaces text prompts with learnable normal/abnormal visual tokens, and UniADet~\cite{gao2026one} learns task- and level-specific visual decision weights for universal anomaly classification and segmentation. Although these methods demonstrate the promise of prompt-free anomaly modeling, their anomaly references are still largely shared across images or tasks. Such static visual anchors can be poorly calibrated under cross-domain medical shifts, where normal appearance and abnormal evidence depend strongly on anatomy, modality, and local context. Different from these methods, ReCAP learns language-free visual prototypes that are conditionally re-centered for each input and further complemented by support-derived normal references for few-shot medical anomaly detection.

\section{Methodology}
\label{sec:method}
\subsection{Problem Formulation}

We study language-free medical anomaly detection under zero-shot and few-shot settings. Given a query image $x\in\mathbb{R}^{H\times W\times3}$, the goal is to predict an image-level anomaly score $s(x)$ and, when pixel-level annotations are available, an anomaly map $\mathbf{A}(x)\in\mathbb{R}^{H\times W}$. The benchmark contains multiple medical domains $\{\mathcal{D}^{(j)}\}_{j=1}^{J}$. In the leave-one-out zero-shot setting, one domain is held out for testing, while the remaining domains form $\mathcal{D}_{\mathrm{src}}$ for source-domain training. In the few-shot setting, a target-domain support set $\mathcal{S}=\{(x_i,y_i,m_i)\}_{i=1}^{N}$ is provided, where $y_i\in\{0,1\}$ denotes the image-level label and $m_i$ is the lesion mask when available.

ReCAP performs anomaly scoring entirely in the visual feature space using two complementary branches (Figure~\ref{fig:framework}). The conditional prototype branch produces an input-specific normal/abnormal boundary, while the few-shot memory branch measures deviations from normal support references. The complete procedure is provided in Appendix~\ref{app:detailed_algorithm}.

\subsection{Multi-level Features and Visual Prototypes}
\paragraph{Multi-level Patch Tokens.}
We use a frozen CLIP visual encoder and insert lightweight segmentation and detection adapters into selected transformer layers $\Omega$ ($\{6,12,18,24\}$ for ViT-L/14). The segmentation adapter preserves spatially faithful patch features for dense anomaly localization, while the detection adapter produces discriminative patch responses for image-level anomaly scoring. We extract $\ell_2$-normalized multi-level patch tokens from both adapters and use them for prototype-based scoring. Detailed multi-level patch token extraction is provided in Appendix~\ref{app:multilevel_features}.

\paragraph{Visual Prototypes.}
In place of text embeddings, each branch and layer maintains a normal prototype $q_l^{b,n}\in\mathbb{R}^{C}$ and abnormal prototype $q_l^{b,a}\in\mathbb{R}^{C}$. These prototypes define a base normal/abnormal boundary that is re-centered for each input by the conditional modulation module. In the few-shot setting, the prototypes are initialized from the corresponding normal and abnormal support patch statistics; in the zero-shot setting, source-trained prototypes serve as the base anchors for the unseen target. In both settings, the prototypes are optimized together with the adapters and conditioning modules. Detailed initialization is provided in Appendix~\ref{app:prototype_initialization}.

\subsection{Conditional Prototype Modulation}
A static prototype is a single point in feature space, yet the optimal normal/abnormal boundary depends on the location of the input. For example, the normal-feature distribution of brain MRI sits far from that of liver CT or retinal OCT. We therefore make the boundary input-conditional and predict it for each image based on its own appearance.

\paragraph{Input Context Descriptor.}
For each branch and layer, we summarize the query into a context descriptor by average-pooling its patch tokens followed by normalization:
\begin{equation}
c_l^b(x)=\operatorname{Norm}\left(\frac{1}{R_l}\sum_{r=1}^{R_l} p_{l,r}^b(x)\right)\in\mathbb{R}^{C}.
\end{equation}
This descriptor captures the global appearance of the image, such as its modality and dominant structure, and serves as the conditioning signal.

\paragraph{Bounded Gated Modulation.}
Each prototype is re-centered by the context through a bounded gated residual:
\begin{equation}
\tilde{q}_l^{b,z}
=
\operatorname{Norm}\Big(
q_l^{b,z}
+
\sigma(\beta_l^{b,z})\,\eta\,
\tanh\big(W_l^{b,z}\,c_l^b(x)\big)
\Big),
\end{equation}
where $W_l^{b,z}\in\mathbb{R}^{C\times C}$ and $\beta_l^{b,z}\in\mathbb{R}$ are learnable, $\sigma(\cdot)$ is the sigmoid, and $\eta>0$ is a fixed modulation strength. The bound on this residual is deliberate and specific to anomaly detection. Because $c_l^b(x)$ is pooled over the entire image, on an abnormal input, it is \emph{contaminated} by the lesion itself. An unconstrained modulation could exploit this by shifting the normal anchor toward the anomalous content, thereby lowering the training loss while making genuine anomalies appear normal at test time. To mitigate this, the learnable gate $\sigma(\beta_l^{b,z})\in(0,1)$ and the bounded $\tanh$ keep the residual small, so that the conditioned anchor $\tilde{q}_l^{b,z}$ stays close to the learned base prototype. The boundary, therefore, re-centers toward the current domain rather than collapsing onto arbitrary content. As a result, a single set of source-trained prototypes can adapt to an unseen domain in a single forward pass, providing an amortized, gradient-free alternative to test-time adaptation that requires no target labels.

\paragraph{Patch-level and Image-level Scoring.}
For a query patch $\hat{p}_{l,r}^b(x)$, we form scaled-cosine logits against the conditioned anchors:
\begin{equation}
g_{l,r}^{b}(x) =
\alpha
\big[\,
\langle \hat{p}_{l,r}^{b}(x),\tilde{q}_l^{b,n}\rangle,
\langle \hat{p}_{l,r}^{b}(x),\tilde{q}_l^{b,a}\rangle
\,\big]\in\mathbb{R}^{2},
\end{equation}
with a fixed logit scale $\alpha$. The patch-level abnormal probability is the abnormal entry of the softmax, $a_{l,r}^{b}(x)=\operatorname{softmax}\big(g_{l,r}^{b}(x)\big)_{a}$. Equivalently, $a_{l,r}^{b}(x)=\sigma\big(\phi_{l,r}^{b}(x)\big)$ with the patch anomaly logit:
\begin{equation}
\phi_{l,r}^{b}(x)=\alpha\big(\langle \hat{p}_{l,r}^{b}(x),\tilde{q}_l^{b,a}\rangle-\langle \hat{p}_{l,r}^{b}(x),\tilde{q}_l^{b,n}\rangle\big),
\end{equation}
which is the normal-versus-abnormal margin under the conditioned boundary. For localization, the segmentation adapter reshapes the two-class patch logits $\{g_{l,r}^{\mathrm{seg}}(x)\}$ into 2D logit maps and upsamples them to the input resolution. A channel-wise softmax is then applied, and the abnormal channel forms the layer-wise anomaly map. For image-level detection, the patch-level abnormal probabilities $\{a_{l,r}^{\mathrm{det}}(x)\}$ are aggregated using the multi-layer fusion described next.

\subsection{Normal-Reference Memory and Layer Fusion}

\paragraph{Normal-Reference Memory.}
The conditional prototype branch models normality using a compact re-centered anchor. To retain instance-level normal variation in the few-shot setting, we build a non-parametric memory $\mathcal{M}_l^b$ from the patch tokens of all normal support images ($y_i=0$). Each query patch is scored by its nearest-neighbor cosine distance:
\begin{equation}
d_{l,r}^b(x)
=
\min_{u\in\mathcal{M}_l^b}
\left(
1-\langle \hat{p}_{l,r}^b(x),\hat{u}\rangle
\right),
\end{equation}
where $\hat{u}$ is an $\ell_2$-normalized memory feature. Since the cosine distance between normalized features lies in $[0,2]$, we map it to the prototype probability range as
\begin{equation}
\bar{d}_{l,r}^{b}(x)
=
\operatorname{clip}
\left(
\frac{d_{l,r}^{b}(x)}{2},\,0,\,1
\right),
\end{equation}
where $\bar{d}_{l,r}^{b}(x)$ denotes the normalized memory anomaly response used for subsequent layer and branch fusion. Large normalized distances indicate deviations from observed target-domain normal patterns. The memory branch therefore complements the compact prototype boundary with instance-level normal references.
\paragraph{Learnable Layer Fusion.}
Both branches produce anomaly evidence at each selected layer in $\Omega$. For image-level detection, the patch-level abnormal probabilities are aggregated by top-$k$ pooling and fused with learnable layer weights:
\begin{equation}
\begin{aligned}
s_{\mathrm{tok}}(x)
&=
\sum_{l\in\Omega}
\omega_l
\operatorname{TopK}_{r}
\left(a_{l,r}^{\mathrm{det}}(x)\right),\\
\{\omega_l\}_{l\in\Omega}
&=
\operatorname{softmax}(\theta),
\end{aligned}
\end{equation}
where $\operatorname{TopK}_{r}(\cdot)$ averages the largest $k$ abnormal probabilities within each layer. The memory score is aggregated from the normalized memory distances using the same layer weights:
\begin{equation}
s_{\mathrm{mem}}(x)
=
\sum_{l\in\Omega}
\omega_l
\operatorname{TopK}_{r}
\left(\bar{d}_{l,r}^{\mathrm{det}}(x)\right).
\end{equation}
For localization, the patch-level abnormal probabilities and normalized memory distances from the segmentation branch are reshaped into 2D maps, upsampled to the input resolution, and fused with segmentation-specific layer weights to obtain $A_{\mathrm{tok}}(x)$ and $A_{\mathrm{mem}}(x)$.

\paragraph{Branch Fusion and Inference.}

In the few-shot setting, the prototype and memory predictions are fused as
\begin{equation}
s(x)
=
(1-\lambda)s_{\mathrm{tok}}(x)
+\lambda s_{\mathrm{mem}}(x),
\end{equation}
where $\lambda\in[0,1]$ balances the two branches and is set to $0.5$ by default. In the zero-shot setting, no memory is constructed, and prediction reduces to the prototype branch with $\lambda=0$. The pixel-level anomaly map $A(x)$ is fused analogously from $A_{\mathrm{tok}}(x)$ and $A_{\mathrm{mem}}(x)$.

\subsection{Training Objectives}
ReCAP is optimized on the source domains $\mathcal{D}_{\mathrm{src}}$ in the zero-shot setting or on the target-domain support set $\mathcal{S}$ in the few-shot setting. The training objective contains three terms: a pixel-level localization loss when lesion masks are available, an image-level detection loss, and a prototype separation regularizer. The first two losses supervise the task-specific scoring functions, while the separation regularizer keeps the base normal/abnormal prototypes discriminative before input-specific re-centering.

\paragraph{Pixel-level Localization and Image-level Detection.}
For a training sample $(x_i,y_i,m_i)$, when the lesion mask $m_i$ is available, the predicted anomaly map $\mathbf{A}(x_i)$ is supervised by a focal-plus-Dice loss:
\begin{equation}
\mathcal{L}_{\mathrm{seg}}
=
\mathcal{L}_{\mathrm{focal}}\big(\mathbf{A}(x_i),m_i\big)
+
\mathcal{L}_{\mathrm{dice}}\big(\mathbf{A}(x_i),m_i\big).
\end{equation}
This loss encourages high anomaly responses on lesion regions and suppresses activations on normal tissues. For image-level detection, the anomaly score $s(x_i)$ is supervised by binary cross-entropy:
\begin{equation}
\mathcal{L}_{\mathrm{det}}
=
\operatorname{BCEWithLogits}
\left(
\operatorname{logit}(s(x_i)),y_i
\right).
\end{equation}
This objective trains the detection branch to separate normal and abnormal images under the same scoring function used at inference.

\paragraph{Prototype Separation Regularization.}
Since conditional modulation re-centers an existing normal/abnormal boundary, the base prototypes should remain discriminative before adaptation. We therefore enforce a margin between the normal and abnormal prototypes in each branch and layer:
\begin{equation}
\mathcal{L}_{\mathrm{sep}}
=
\frac{1}{2|\Omega|}
\sum_{b\in\{\mathrm{seg},\mathrm{det}\}}
\sum_{l\in\Omega}
\max\big(0,\cos(q_l^{b,n},q_l^{b,a})-\delta\big),
\end{equation}
where $\delta$ is the separation margin. This regularizer is coupled with bounded modulation: the margin preserves a discriminative base boundary, while the modulation bound allows the boundary to shift toward the current input without collapsing onto lesion-contaminated context.

\paragraph{Overall Loss Function.}
The final training objective is:
\begin{equation}
\mathcal{L}
=
\mathcal{L}_{\mathrm{seg}}
+
\mathcal{L}_{\mathrm{det}}
+
\lambda_{\mathrm{sep}}\mathcal{L}_{\mathrm{sep}}.
\end{equation}
For datasets without pixel-level annotations, $\mathcal{L}_{\mathrm{seg}}$ is omitted, and ReCAP is trained with image-level supervision and prototype separation.

\begin{table*}[!t]
\centering
\scriptsize
\setlength{\tabcolsep}{2.8pt}
\resizebox{\textwidth}{!}{%
\begin{tabular}{lclccccccc}
\toprule
\multirow{2}{*}{\textbf{Few-shot}} &
\multirow{2}{*}{\textbf{Metric}} &
\multirow{2}{*}{\textbf{Dataset}} &
\textbf{DRA} &
\textbf{APRIL-GAN} &
\textbf{MediCLIP} &
\textbf{MVFA} &
\textbf{MadCLIP} &
\textbf{VisualAD} &
\cellcolor{oursblue}\textbf{ReCAP} \tabularnewline
\cmidrule(lr){4-10}
& & &
N/A &
\xmark &
\xmark &
\xmark &
\xmark &
\cmark &
\cellcolor{oursblue}\cmark \tabularnewline
\midrule

\multirow{9}{*}{$K{=}2$}
& \multirow{6}{*}{\makecell[c]{Image-level\\(AUROC, $F_1$-max, AP)}}
& HIS
& (72.91, 71.52, 73.20)
& (69.57, 69.24, 70.52)
& (64.49, 66.81, 65.20)
& (79.75, 76.96, 71.72)
& (\underline{83.62}, \underline{77.60}, \underline{84.22})
& (53.80, 67.13, 53.14)
& \cellcolor{oursblue}(\textbf{85.89}, \textbf{78.44}, \textbf{86.10}) \tabularnewline

& & Chest
& (72.22, 96.82, 97.82)
& (69.84, 96.64, 97.48)
& (61.69, 96.39, 96.12)
& (78.38, \underline{97.68}, 98.48)
& (\underline{84.56}, 97.60, \underline{98.90})
& (53.70, \textbf{97.70}, 95.62)
& \cellcolor{oursblue}(\textbf{84.71}, \underline{97.68}, \textbf{99.00}) \tabularnewline

& & OCT17
& (98.08, 96.41, 98.70)
& (\underline{99.21}, 92.80, \underline{99.40})
& (93.40, 92.73, 96.10)
& (96.53, 96.83, 98.56)
& (99.06, \underline{97.50}, 99.30)
& (92.40, 91.02, 97.64)
& \cellcolor{oursblue}(\textbf{99.40}, \textbf{98.26}, \textbf{99.70}) \tabularnewline

& & BrainMRI
& (71.78, 84.64, 89.20)
& (78.45, 88.34, 92.40)
& (85.13, 90.61, 94.80)
& (91.95, 92.88, \underline{98.23})
& (\underline{93.93}, \textbf{94.20}, 98.20)
& (83.90, 92.04, 95.33)
& \cellcolor{oursblue}(\textbf{94.35}, \underline{94.08}, \textbf{98.48}) \tabularnewline

& & LiverCT
& (57.17, 62.34, 65.89)
& (57.80, 63.11, 66.23)
& (68.48, 70.20, 74.51)
& (\underline{87.23}, 77.79, \underline{88.74})
& (84.48, \underline{78.40}, 87.30)
& (84.42, 74.23, 84.91)
& \cellcolor{oursblue}(\textbf{90.45}, \textbf{80.20}, \textbf{90.00}) \tabularnewline

& & RESC
& (85.69, 78.82, 85.60)
& (89.44, 81.24, 88.23)
& (83.96, 78.11, 84.40)
& (94.70, 84.87, \underline{93.58})
& (\underline{95.09}, \underline{86.70}, 93.40)
& (91.20, 82.60, 87.42)
& \cellcolor{oursblue}(\textbf{95.47}, \textbf{87.29}, \textbf{94.33}) \tabularnewline

\cmidrule(lr){2-10}

& \multirow{3}{*}{\makecell[c]{Pixel-level\\(AUROC, $F_1$-max, PRO)}}
& BrainMRI
& (72.09, 18.44, 52.64)
& (94.02, 36.82, 78.40)
& (97.39, 42.60, 86.74)
& (97.41, 35.84, 77.85)
& (\textbf{97.92}, 45.21, \underline{88.64})
& (95.00, \underline{45.71}, 80.21)
& \cellcolor{oursblue}(\underline{97.52}, \textbf{50.24}, \textbf{89.03}) \tabularnewline

& & LiverCT
& (63.13, 20.10, 48.54)
& (95.87, 44.62, 78.22)
& (97.09, 51.83, 82.44)
& (96.67, 62.17, 82.79)
& (99.39, 59.40, \underline{87.60})
& (99.10, 34.90, 76.67)
& \cellcolor{oursblue}(\textbf{99.64}, \textbf{62.89}, \textbf{84.61}) \tabularnewline

& & RESC
& (65.59, 28.62, 46.81)
& (96.39, 63.50, 75.64)
& (96.01, 61.84, 74.20)
& (97.10, 62.34, \textbf{87.55})
& (\underline{97.18}, \underline{67.80}, 78.40)
& (97.10, 64.90, 78.16)
& \cellcolor{oursblue}(\textbf{98.37}, \textbf{70.85}, \underline{78.91}) \tabularnewline

\midrule

\multirow{9}{*}{$K{=}4$}
& \multirow{6}{*}{\makecell[c]{Image-level\\(AUROC, $F_1$-max, AP)}}
& HIS
& (68.73, 69.84, 68.71)
& (76.11, 73.20, 76.20)
& (70.85, 70.64, 72.83)
& (\underline{80.05}, \underline{77.50}, 75.67)
& (\underline{80.05}, 76.40, \underline{78.26})
& (67.30, 70.00, 64.40)
& \cellcolor{oursblue}(\textbf{82.29}, \textbf{79.69}, \textbf{79.85}) \tabularnewline

& & Chest
& (75.81, 96.30, 98.10)
& (77.43, 97.14, 98.20)
& (56.83, 96.12, 95.40)
& (82.51, \underline{97.69}, 97.84)
& (\underline{88.14}, 95.70, \textbf{99.20})
& (65.20, 97.30, 97.50)
& \cellcolor{oursblue}(\textbf{88.92}, \textbf{98.68}, \underline{98.84}) \tabularnewline

& & OCT17
& (99.06, 97.34, 99.30)
& (\underline{99.41}, 98.00, 99.45)
& (89.07, 91.50, 95.40)
& (99.38, \underline{98.64}, \underline{99.77})
& (99.37, 98.15, 99.63)
& (98.30, 96.42, 99.43)
& \cellcolor{oursblue}(\textbf{99.86}, \textbf{99.24}, \textbf{99.95}) \tabularnewline

& & BrainMRI
& (80.62, 88.22, 93.44)
& (89.18, 92.30, 96.80)
& (83.82, 90.14, 94.12)
& (88.89, 91.62, 97.57)
& (\textbf{95.25}, \textbf{95.40}, \underline{98.28})
& (77.50, 91.23, 93.62)
& \cellcolor{oursblue}(\underline{95.14}, \underline{95.01}, \textbf{98.90}) \tabularnewline

& & LiverCT
& (59.64, 64.12, 68.55)
& (53.05, 60.70, 62.82)
& (81.53, 75.64, 84.74)
& (82.10, 73.49, 83.95)
& (\underline{82.97}, \underline{76.80}, \underline{86.20})
& (59.40, 61.92, 56.63)
& \cellcolor{oursblue}(\textbf{87.56}, \textbf{77.30}, \textbf{87.47}) \tabularnewline

& & RESC
& (90.90, 82.32, 89.44)
& (94.70, 85.40, 93.22)
& (87.52, 80.25, 87.10)
& (94.67, 87.04, 91.61)
& (\underline{96.62}, \underline{88.30}, \underline{95.60})
& (91.20, 81.40, 87.60)
& \cellcolor{oursblue}(\textbf{96.94}, \textbf{89.29}, \textbf{96.62}) \tabularnewline

\cmidrule(lr){2-10}

& \multirow{3}{*}{\makecell[c]{Pixel-level\\(AUROC, $F_1$-max, PRO)}}
& BrainMRI
& (74.77, 20.62, 55.30)
& (94.67, 38.54, 79.82)
& (96.86, 41.80, 84.70)
& (95.05, 36.64, 76.91)
& (\textbf{97.90}, \underline{45.20}, \underline{88.20})
& (84.90, 19.80, 82.10)
& \cellcolor{oursblue}(\underline{97.56}, \textbf{56.76}, \textbf{89.19}) \tabularnewline

& & LiverCT
& (71.79, 25.32, 54.72)
& (96.24, 47.90, 79.44)
& (98.61, 55.44, 84.61)
& (\underline{99.65}, \underline{60.10}, \textbf{91.05})
& (99.29, 58.81, 87.23)
& (98.40, 22.20, 79.22)
& \cellcolor{oursblue}(\textbf{99.77}, \textbf{61.08}, \underline{89.42}) \tabularnewline

& & RESC
& (77.28, 34.22, 55.82)
& (97.98, 70.40, 82.60)
& (96.65, 64.54, 76.84)
& (98.73, \underline{77.86}, \textbf{93.42})
& (\underline{98.90}, 75.80, 89.80)
& (97.50, 65.00, 90.40)
& \cellcolor{oursblue}(\textbf{99.11}, \textbf{78.05}, \underline{92.37}) \tabularnewline

\bottomrule
\end{tabular}%
}
\caption{Comparisons with state-of-the-art few-shot anomaly detection methods under different few-shot settings ($K{=}2,4$). Image-level results are reported as (AUROC, $F_1$-max, AP), and pixel-level results as (AUROC, $F_1$-max, PRO). \cmark denotes language-free methods, while \xmark denotes methods using text prompts, and N/A denotes non-CLIP methods. All values are reported in percentage.}
\label{tab:sota_fewshot_k2_k4_full_metrics}
\end{table*}

\section{Experiments}

\subsection{Experimental Setup}
\paragraph{Datasets.}
Following prior medical anomaly detection benchmarks~\cite{huang2024adapting}, we evaluate ReCAP on six datasets spanning five imaging domains. These include BrainMRI for brain MRI~\cite{RN21,bakas2017advancing,menze2014multimodal}, LiverCT for liver CT~\cite{bilic2023liver,landman2015miccai}, RESC and OCT17 for retinal OCT~\cite{kermany2018identifying,hu2019automated}, ChestXray for chest X-ray~\cite{wang2017chestx}, and HIS for digital histopathology~\cite{ehteshami2017diagnostic}. BrainMRI, LiverCT, and RESC contain pixel-level annotations and are evaluated for both anomaly classification (AC) and anomaly segmentation (AS); the remaining datasets are evaluated for AC only. Dataset statistics and preprocessing details are provided in Appendix~\ref{app:dataset_details}.

\paragraph{Competing Methods and Baselines.}
We compare ReCAP with representative medical anomaly detection methods, including DRA~\cite{ding2022catching}, APRIL-GAN~\cite{chen2023zero}, MediCLIP~\cite{zhang2024mediclip}, MVFA~\cite{huang2024adapting}, MadCLIP~\cite{shiri2025madclip}, and VisualAD~\cite{hou2026visualad}. For fair comparison, all methods are evaluated under the same dataset splits, support-shot settings, and image-level or pixel-level evaluation protocols whenever applicable. For all few-shot experiments, every method uses the same predefined support indices for each dataset and shot setting.

\paragraph{Evaluation Metrics.}
For anomaly classification (AC), we report image-level AUROC, the maximum $F_1$ score across decision thresholds ($F_1$-max), and average precision (AP). For anomaly segmentation (AS), we use pixel-level AUROC, pixel-level $F_1$-max, and per-region overlap (PRO). All metrics are reported as percentages.

\paragraph{Implementation Details.}
ReCAP uses a frozen CLIP ViT-L/14@336px visual encoder with lightweight segmentation and detection adapters. Images are resized to $240\times240$, and patch tokens from layers $\{6,12,18,24\}$ are used for multi-level anomaly scoring. Only the adapters, visual prototypes, conditioning modules, and layer-fusion weights are trainable. Training combines image-level detection, pixel-level localization when masks are available, and prototype separation. Complete optimization settings, support protocols, augmentation, and computing resources are listed in Appendix~\ref{app:implementation_details}.

\subsection{Comparison with State-of-the-Art Methods}

\begin{table*}[t]
\centering
\scriptsize
\setlength{\tabcolsep}{3.0pt}
\resizebox{\textwidth}{!}{%
\begin{tabular}{lclcccccc}
\toprule
\multirow{2}{*}{\textbf{Setting}} &
\multirow{2}{*}{\textbf{Metric}} &
\multirow{2}{*}{\textbf{Dataset}} &
\textbf{APRIL-GAN} &
\textbf{MediCLIP} &
\textbf{MVFA} &
\textbf{MadCLIP} &
\textbf{VisualAD} &
\cellcolor{oursblue}\textbf{ReCAP} \tabularnewline
\cmidrule(lr){4-9}
& & &
\xmark &
\xmark &
\xmark &
\xmark &
\cmark &
\cellcolor{oursblue}\cmark \tabularnewline
\midrule

\multirow{9}{*}{Zero-shot}
& \multirow{6}{*}{\makecell[c]{Image-level\\(AUROC, $F_1$-max, AP)}}
& HIS
& (59.36, 66.21, 56.72)
& (60.60, 64.20, 62.31)
& (76.89, 73.73, 70.95)
& (\underline{77.32}, \underline{74.80}, \underline{80.16})
& (50.50, 64.21, 50.82)
& \cellcolor{oursblue}(\textbf{79.32}, \textbf{75.86}, \textbf{82.41}) \tabularnewline

& & Chest
& (57.49, 95.20, 95.84)
& (58.20, 94.90, 94.85)
& (71.11, \underline{96.42}, 97.19)
& (\underline{75.26}, 96.85, \underline{97.94})
& (45.30, 96.40, 94.31)
& \cellcolor{oursblue}(\textbf{78.40}, \textbf{96.92}, \textbf{98.21}) \tabularnewline

& & OCT17
& (92.61, 89.69, 96.97)
& (87.89, 88.72, 93.43)
& (\underline{95.40}, 93.52, \underline{98.12})
& (91.48, 90.80, 96.32)
& (88.37, 88.20, 96.62)
& \cellcolor{oursblue}(\textbf{96.58}, \textbf{95.73}, \textbf{98.84}) \tabularnewline

& & BrainMRI
& (76.63, 86.40, 90.35)
& (\underline{79.32}, 87.92, 91.61)
& (79.80, \textbf{92.18}, \underline{93.78})
& (74.43, 90.50, 93.06)
& (72.36, 83.28, 86.98)
& \cellcolor{oursblue}(\textbf{80.24}, \underline{91.05}, \textbf{94.20}) \tabularnewline

& & LiverCT
& (54.62, 60.21, 63.45)
& (62.13, 67.64, 70.25)
& (\underline{81.19}, 72.27, 76.94)
& (73.48, \textbf{74.92}, \underline{82.40})
& (70.20, 69.32, 80.15)
& \cellcolor{oursblue}(\textbf{82.60}, \underline{73.64}, \textbf{83.43}) \tabularnewline

& & RESC
& (75.67, 76.12, 80.35)
& (80.50, 74.90, 82.15)
& (\underline{89.00}, 80.48, 84.12)
& (84.10, \underline{82.35}, 88.40)
& (81.70, 78.12, 84.23)
& \cellcolor{oursblue}(\textbf{90.23}, \textbf{82.78}, \textbf{88.95}) \tabularnewline

\cmidrule(lr){2-9}

& \multirow{3}{*}{\makecell[c]{Pixel-level\\(AUROC, $F_1$-max, PRO)}}
& BrainMRI
& (\underline{91.79}, 28.50, 69.42)
& (90.21, 30.80, 70.20)
& (89.68, 14.49, 59.26)
& (91.40, \underline{32.60}, \underline{74.50})
& (80.40, 12.89, 67.21)
& \cellcolor{oursblue}(\textbf{92.30}, \textbf{36.85}, \textbf{79.20}) \tabularnewline

& & LiverCT
& (93.05, 32.10, 66.45)
& (94.72, 38.65, 71.30)
& (93.93, 18.94, 65.46)
& (\underline{98.30}, \underline{43.50}, \underline{75.20})
& (97.20, 30.12, 69.40)
& \cellcolor{oursblue}(\textbf{98.72}, \textbf{58.32}, \textbf{78.46}) \tabularnewline

& & RESC
& (85.23, 52.12, 63.42)
& (89.62, 54.82, 65.30)
& (\underline{90.44}, 35.99, 61.42)
& (90.10, \underline{58.46}, \underline{68.70})
& (84.30, 50.24, 62.60)
& \cellcolor{oursblue}(\textbf{92.60}, \textbf{60.42}, \textbf{70.35}) \tabularnewline

\bottomrule
\end{tabular}%
}
\caption{Comparisons with state-of-the-art zero-shot anomaly detection methods. Image-level results are reported as (AUROC, $F_1$-max, AP), and pixel-level results as (AUROC, $F_1$-max, PRO). \cmark denotes language-free methods, while \xmark denotes methods using text prompts. All values are reported in percentage.}
\label{tab:sota_zeroshot_all}
\end{table*}

\paragraph{Quantitative Comparison.}
Table~\ref{tab:sota_fewshot_k2_k4_full_metrics} compares ReCAP with representative few-shot anomaly detection methods under low-shot settings ($K=2,4$). Under $K=2$, ReCAP achieves the best image-level AUROC on all six datasets. In particular, it improves the LiverCT AUROC from $84.48\%$ with MadCLIP to $90.45\%$. For pixel-level segmentation, ReCAP achieves the best AUROC on LiverCT and RESC, reaching $99.64\%$ and $98.37\%$, respectively, while obtaining the best $F_1$-max on all three segmentation datasets. Under $K=4$, ReCAP maintains the best image-level AUROC on five of six datasets and the best pixel-level AUROC on LiverCT and RESC. It also achieves the best pixel-level $F_1$-max on all three datasets. These results demonstrate the effectiveness of ReCAP for both anomaly classification and localization with limited target-domain supervision.

Table~\ref{tab:sota_zeroshot_all} reports zero-shot results on six medical datasets. Without using target-domain support samples, ReCAP achieves the best image-level AUROC on all six datasets, outperforming both prompt-based CLIP methods and the language-free VisualAD baseline. Notable improvements include an increase from $75.26\%$ with MadCLIP to $78.40\%$ on Chest and from $95.40\%$ with MVFA to $96.58\%$ on OCT17. For pixel-level segmentation, ReCAP achieves the best AUROC and $F_1$-max on all three segmentation datasets. These results demonstrate that the proposed language-free conditional prototypes generalize effectively to unseen medical domains while retaining accurate anomaly localization. Appendix~\ref{app:additional_quantitative} further reports results for $K=8,16$ and provides mean$\pm$std AUROC over five random seeds for all shot settings ($K=2,4,8,16$), showing that the improvements are robust to random variation.

\begin{figure}[ht]
    \centering
    \includegraphics[width=\linewidth]{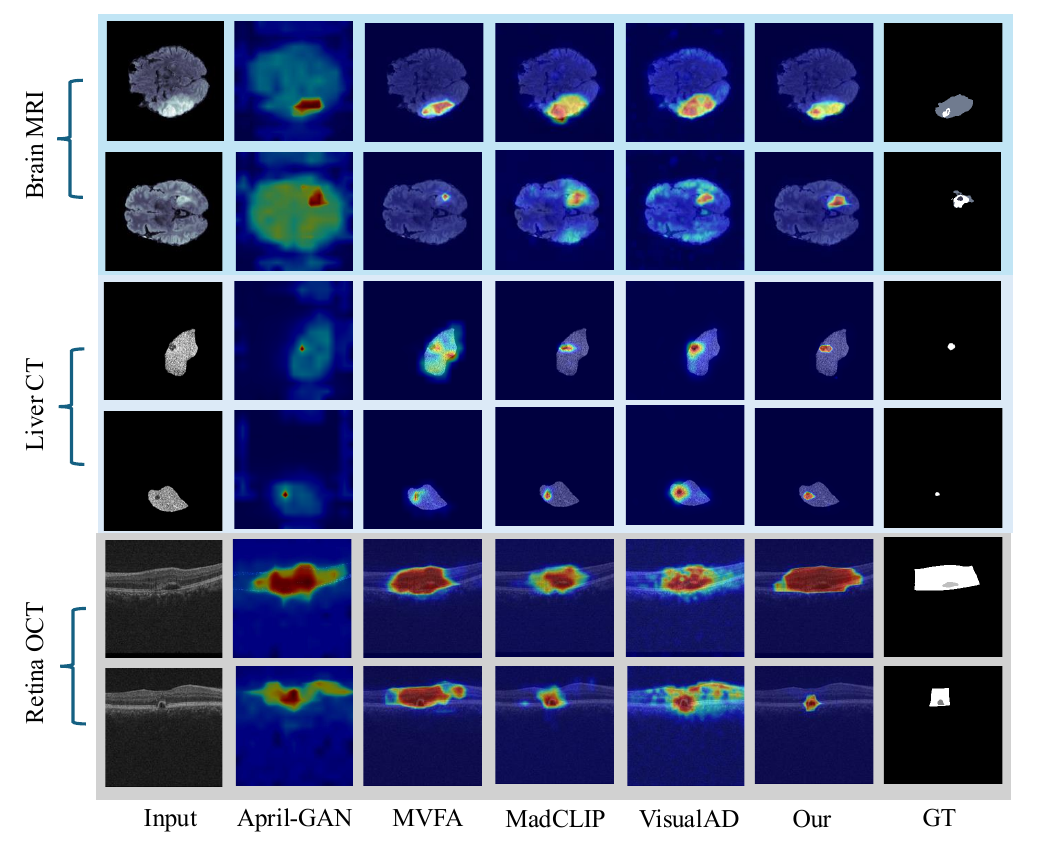}
    \caption{Qualitative comparison of anomaly localization. We compare heatmaps produced by APRIL-GAN, MVFA, MadCLIP, VisualAD, and ReCAP on representative samples from BrainMRI, LiverCT, and retinal OCT datasets.}
    \label{fig:qualitative_comparison}
\end{figure}

\paragraph{Qualitative Comparison.}
Figure~\ref{fig:qualitative_comparison} compares anomaly localization results on BrainMRI, LiverCT, and retinal OCT images. APRIL-GAN produces diffuse heatmaps that extend substantially into normal regions. MVFA and MadCLIP yield more localized responses but occasionally show incomplete lesion coverage or off-target activations, while VisualAD incorrectly highlights normal anatomical structures in several BrainMRI examples. In contrast, ReCAP produces more compact anomaly responses that are visually better aligned with the annotated lesions and exhibit fewer background activations across the three modalities. These results qualitatively support the localization performance reported in our quantitative evaluation. Additional visualizations are provided in Appendix~\ref{app:additional_visualizations}.

\begin{figure}[t]
    \centering
    \includegraphics[width=\linewidth]{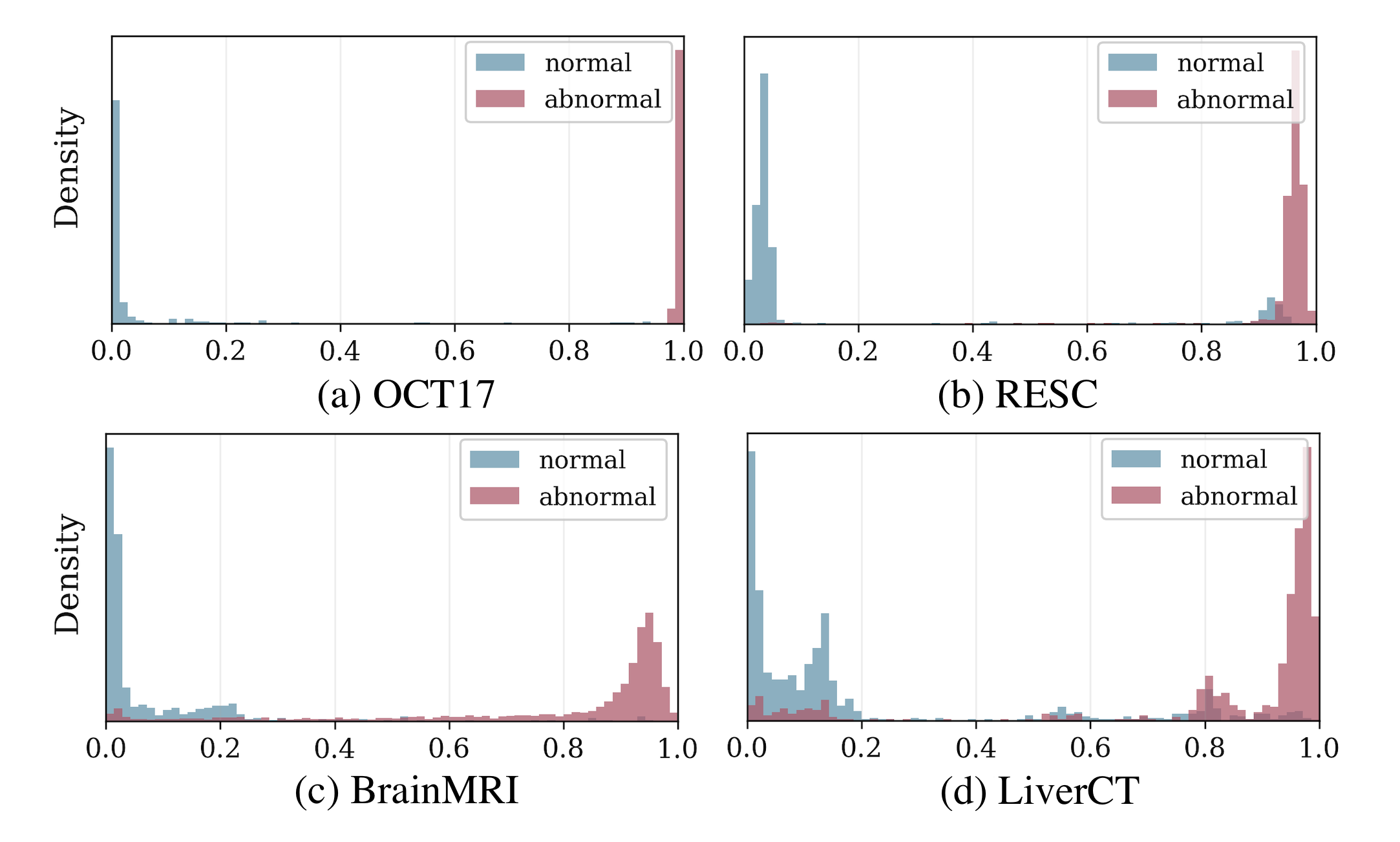}
    \caption{Image-level anomaly score distributions under the 16-shot setting. Normalized scores for normal (blue) and abnormal (red) test samples on four representative datasets.}
    \label{fig:score_distribution}
\end{figure}

\paragraph{Anomaly Score Distribution Analysis.}
Figure~\ref{fig:score_distribution} shows representative image-level score distributions on OCT17, RESC, BrainMRI, and LiverCT, with the other two datasets reported in Appendix~\ref{app:additional_visualizations}. ReCAP assigns low scores to most normal samples and high scores to abnormal samples. The separation is particularly clear on OCT17, RESC, and BrainMRI, demonstrating well-separated anomaly responses across retinal and brain imaging domains. Although the distributions partially overlap on the more challenging LiverCT dataset, abnormal samples remain concentrated toward higher scores.

\paragraph{Efficiency Comparison.}
Table~\ref{tab:complexity_efficiency} compares model complexity and inference latency under the same input resolution, hardware, and measurement protocol. ReCAP achieves the fastest inference in both zero-shot and four-shot settings, requiring only \textbf{17.4} ms and \textbf{19.0} ms per image, respectively. The comparison with VisualAD is especially informative because both methods are language-free and use the same ViT-L/14@336px backbone. ReCAP is approximately $3.8\times$ faster in zero-shot inference ($17.4$ vs.\ $66.1$ ms) and $3.5\times$ faster in the four-shot setting ($19.0$ vs.\ $65.8$ ms), while using fewer method-specific parameters ($22.0$M vs.\ $31.5$M). This efficiency gain stems from the different inference designs: VisualAD propagates learnable normal and abnormal tokens through multi-layer token--patch interactions and additional spatial recalibration modules, whereas ReCAP performs lightweight bounded prototype modulation followed by direct patch--prototype cosine scoring. Compared with prompt-based methods, ReCAP further avoids text encoding and image--text matching. This makes ReCAP more suitable for time-sensitive clinical screening workflows. Additional analyses of the visual backbone, fusion weight $\lambda$, and top-$k$ ratio are provided in Appendix~\ref{app:parameter_analysis}.

\begin{table}[!ht]
\centering
\resizebox{\linewidth}{!}{
\begin{tabular}{c l c c c c}
\toprule
Shots & Methods & CLIP Models & Input Size & \# Params (M) & Inf. Time (ms) \\
\midrule
0 & APRIL-GAN & ViT-B/16 & $240\times240$ & 149.6 + 1.2 & 114.5 $\pm$ 3.7 \\
0 & MVFA & ViT-L/14 & $240\times240$ & 427.7 + 12.6 & 125.7 $\pm$ 5.3 \\
0 & MadCLIP & ViT-L/14 & $240\times240$ & 427.7 + 106.9 & 265.3 $\pm$ 11.0 \\
0 & VisualAD & ViT-L/14@336px & $240\times240$ & 427.2 + 31.5 & 66.1 $\pm$ 0.9 \\
0 & ReCAP & ViT-L/14@336px & $240\times240$ & 427.7 + 22.0 & \textbf{17.4 $\pm$ 2.0} \\
\midrule
4 & APRIL-GAN & ViT-B/16 & $240\times240$ & 152.1 + 4.6 & 120.1 $\pm$ 2.4 \\
4 & MVFA & ViT-L/14 & $240\times240$ & 427.7 + 32.3 & 126.2 $\pm$ 4.2 \\
4 & MadCLIP & ViT-L/14 & $240\times240$ & 427.7 + 92.3 & 268.5 $\pm$ 9.3 \\
4 & VisualAD & ViT-L/14@336px & $240\times240$ & 427.2 + 31.5 & 65.8 $\pm$ 0.9 \\
4 & ReCAP & ViT-L/14@336px & $240\times240$ & 427.7  + 22.0 & \textbf{19.0 $\pm$ 4.2} \\
\bottomrule
\end{tabular}
}
\caption{Complexity and inference efficiency under zero- and four-shot settings. Parameter counts are reported as pretrained backbone parameters $+$ additional method-specific parameters. Inference time is reported as mean $\pm$ standard deviation over 100 runs after 20 warm-up iterations.}
\label{tab:complexity_efficiency}
\end{table}

\subsection{Ablation Studies}
\paragraph{Component Analysis.}
Table~\ref{tab:ablation_components} evaluates the contributions of the normal memory bank (MB), context-conditioning module (CCM), and learnable layer fusion (LF). The full model achieves the best AC and AS on both datasets, indicating that all three components contribute to the final performance. On LiverCT, removing LF produces the largest reductions, decreasing AC and AS by $4.00$ and $2.89$ percentage points, respectively, which demonstrates the importance of aggregating anomaly evidence across multiple layers. Removing MB reduces AC and AS by $2.36$ and $2.72$ points, while removing CCM results in drops of $1.60$ and $1.48$ points. On RESC, every component removal also degrades both metrics, although the margins are smaller because performance is already close to saturation. Overall, the results support the joint use of target-domain normal references, context-conditioned prototypes, and multi-layer score fusion.

\begin{table}[!ht]
\centering
\setlength{\tabcolsep}{5pt}
\renewcommand{\arraystretch}{1.12}
\resizebox{\linewidth}{!}{
\begin{tabular}{lccc cc cc}
\toprule
\multirow{2}{*}{Variant} &
\multirow{2}{*}{MB} &
\multirow{2}{*}{CCM} &
\multirow{2}{*}{LF} &
\multicolumn{2}{c}{LiverCT} &
\multicolumn{2}{c}{RESC} \\
\cmidrule(lr){5-6} \cmidrule(lr){7-8}
& & & & AC & AS & AC & AS \\
\midrule
w/o MB   &            & \checkmark & \checkmark & 85.20 & 97.05 & 95.54 & 98.82 \\
w/o CCM  & \checkmark &            & \checkmark & 85.96 & 98.29 & 96.84 & 98.94 \\
w/o LF   & \checkmark & \checkmark &            & 83.56 & 96.88 & 96.91 & 98.97 \\
ReCAP & \checkmark & \checkmark & \checkmark & \textbf{87.56} & \textbf{99.77} & \textbf{96.94} & \textbf{99.11} \\
\bottomrule
\end{tabular}
}
\caption{Component ablation under the 4-shot setting. AC and AS are reported as AUROC (\%).}
\label{tab:ablation_components}
\end{table}

\paragraph{Effect of Controlled Prototype Conditioning.}
Table~\ref{tab:conditioning_ablation} compares different language-free prototype designs. Static prototypes outperform the VisualAD-style anchors on all eight metrics, showing the benefit of ReCAP's normal/abnormal prototype representation even without query conditioning. However, these prototypes remain identical for all query images and cannot adapt to anatomy-dependent appearance variations. Adding an unbounded context residual decreases all four zero-shot metrics and produces mixed results under the 4-shot setting, indicating that unconstrained modulation does not reliably improve the anomaly reference. Bounding the residual improves six of eight metrics over its unbounded counterpart, including all AC and RESC results. Introducing the gate further improves every metric, with the bounded gated residual achieving the best AC and AS across both datasets and shot settings. These results support the use of controlled query-conditioned modulation rather than either static or unconstrained anomaly prototypes.

\begin{table}[t]
\centering
\resizebox{\linewidth}{!}{
\begin{tabular}{lcccccccc}
\toprule
\multirow{3}{*}{Variant}
& \multicolumn{4}{c}{Zero-shot}
& \multicolumn{4}{c}{4-shot} \\
\cmidrule(lr){2-5}
\cmidrule(lr){6-9}
& \multicolumn{2}{c}{LiverCT}
& \multicolumn{2}{c}{RESC}
& \multicolumn{2}{c}{LiverCT}
& \multicolumn{2}{c}{RESC} \\
\cmidrule(lr){2-3}
\cmidrule(lr){4-5}
\cmidrule(lr){6-7}
\cmidrule(lr){8-9}
& AC & AS & AC & AS & AC & AS & AC & AS \\
\midrule
VisualAD-style anchors
& 68.92 & 96.84
& 80.35 & 82.58
& 58.64 & 97.42
& 90.82 & 96.85 \\

Static prototypes
& 70.06 & 98.19
& 84.77 & 90.67
& 80.71 & 98.91
& 95.46 & 97.42 \\

Unbounded residual
& 67.43 & 97.99
& 81.83 & 88.41
& 77.70 & 98.98
& 95.77 & 96.93 \\

Bounded residual w/o gate
& 70.51 & 97.61
& 86.84 & 91.35
& 81.42 & 98.92
& 96.12 & 98.21 \\

Bounded gated residual
& \textbf{82.60} & \textbf{98.72}
& \textbf{90.23} & \textbf{92.60}
& \textbf{87.56} & \textbf{99.77}
& \textbf{96.94} & \textbf{99.11} \\
\bottomrule
\end{tabular}
}
\caption{Ablation of prototype conditioning under zero- and 4-shot settings. The first two variants omit query conditioning: VisualAD-style uses shared learnable anchors, whereas Static prototypes use ReCAP's normal/abnormal prototypes.}
\label{tab:conditioning_ablation}
\end{table}

\section{Conclusion}
This paper presented ReCAP, a language-free framework for zero- and few-shot medical anomaly detection. ReCAP replaces static visual anchors with input-conditioned normal/abnormal prototypes whose updates are bounded and gated to provide adaptive anomaly references. For few-shot transfer, it further combines a normal-reference memory with learnable multi-layer fusion to capture target-domain normal variation and aggregate complementary anomaly cues. Across six medical datasets, ReCAP achieves the best zero-shot image-level AUROC on all six datasets and pixel-level AUROC on all three segmentation datasets, while maintaining strong performance across few-shot settings. Ablation studies confirm the contributions of controlled prototype conditioning, normal-reference memory, and layer fusion, and the efficiency analysis demonstrates fast inference without text prompts. These results establish controlled visual prototype adaptation as an effective alternative to text-based anomaly modeling. Current evaluation focuses on 2D image-based benchmarks; extending ReCAP to volumetric and multimodal clinical data remains an important direction.

\bibliography{recap}

\clearpage
\appendix
\setcounter{equation}{0}
\renewcommand{\theequation}{A\arabic{equation}}
\input{appendix}

\end{document}

%% file: appendix.tex
\section{Detailed Algorithm of ReCAP}
\label{app:detailed_algorithm}

\begin{algorithm}[!ht]
\caption{Training and inference procedures of ReCAP}
\label{alg:recap}

\KwIn{Source domains $\mathcal{D}_{\mathrm{src}}$, optional support set $\mathcal{S}$, query image $x$, frozen encoder $E_v$, layers $\Omega$, and conditioning strength $\eta$.}
\KwOut{Anomaly map $A(x)$ and image-level score $s(x)$.}

Freeze $E_v$ and initialize adapters, prototypes, conditioning modules, and fusion weights.\;

\BlankLine
\textbf{Training:}\;
\eIf{$\mathcal{S}\neq\varnothing$}{
    Initialize support prototypes and normal memory banks from $\mathcal{S}$.\;
    Optimize ReCAP on $\mathcal{S}$ with Eqs.~(10)--(13).\;
}{
    Optimize ReCAP on $\mathcal{D}_{\mathrm{src}}$ with Eqs.~(10)--(13).\;
}

\BlankLine
\textbf{Inference:}\;
Extract multi-level patch tokens from $x$ and generate conditioned prototypes by Eqs.~(1)--(2).\;
Compute prototype responses and layer-fused scores by Eqs.~(3), (4), and (7).\;

\eIf{$\mathcal{S}\neq\varnothing$}{
    Compute memory responses by Eqs.~(5)--(6) and fuse both branches by Eqs.~(8)--(9).\;
}{
    Set $\lambda=0$ and use the prototype branch only.\;
}

\Return{$A(x)$ and $s(x)$.}
\end{algorithm}
Algorithm~\ref{alg:recap} summarizes ReCAP in both zero-shot and few-shot settings. In zero-shot inference, ReCAP uses source-trained visual prototypes and input-conditioned modulation to produce anomaly predictions without target-domain support. In few-shot inference, support images are additionally used to initialize target-domain prototypes and build a normal-reference memory. In both cases, query-specific adaptation is performed by bounded gated prototype conditioning in a single forward pass, without text prompts or test-time gradient updates.

\section{Additional Method Details}
\label{app:method_details}

\subsection{Multi-level Patch Token Extraction}
\label{app:multilevel_features}
ReCAP extracts multi-level patch representations from a frozen CLIP visual encoder with lightweight task-specific adapters. Let $\Omega=\{l_1,\ldots,l_M\}$ denote the selected transformer layers. For each branch $b\in\{\mathrm{seg},\mathrm{det}\}$ and layer $l\in\Omega$, the adapter produces a sequence of patch tokens:
\begin{equation}
P_l^b(x)=
\left[
p_{l,1}^b(x),\ldots,p_{l,R_l}^b(x)
\right]^{\top}
\in \mathbb{R}^{R_l\times C},
\end{equation}
where $R_l$ is the number of patch tokens and $C$ is the feature dimension. We use $\ell_2$-normalized tokens for prototype matching and memory retrieval:
\begin{equation}
\hat{p}_{l,r}^b(x)=
\frac{p_{l,r}^b(x)}
{\lVert p_{l,r}^b(x)\rVert_2}.
\end{equation}
The segmentation branch keeps spatially faithful patch responses for pixel-level anomaly maps, while the detection branch produces patch responses that are aggregated into an image-level anomaly score.

\subsection{Support-based Prototype Initialization}
\label{app:prototype_initialization}
In the few-shot setting, normal and abnormal visual prototypes are initialized from support-set patch statistics. For each branch $b$, layer $l$, and class $z\in\{n,a\}$, we collect the corresponding support patch tokens into $\mathcal{T}_l^{b,z}$. Each prototype is initialized as the normalized mean:
\begin{equation}
q_l^{b,z}
=
\operatorname{Norm}
\left(
\frac{1}{|\mathcal{T}_l^{b,z}|}
\sum_{u\in\mathcal{T}_l^{b,z}} u
\right),
\qquad z\in\{n,a\},
\end{equation}
where $\operatorname{Norm}(v)=v/\lVert v\rVert_2$. The sets $\mathcal{T}_l^{b,n}$ and $\mathcal{T}_l^{b,a}$ contain all patch tokens from normal and abnormal support images, respectively. Lesion masks, when available, are used only for pixel-level supervision and not for prototype initialization. In the zero-shot setting, no target-domain support set is available; source-trained prototypes are directly used as base anchors for the unseen target domain.

\section{Details of Medical Anomaly Detection Datasets}
\label{app:dataset_details}
We evaluate ReCAP on six medical anomaly detection datasets covering diverse modalities, anatomical regions, and evaluation settings. As summarized in Table~\ref{tab:dataset_summary}, BrainMRI, LiverCT, and RESC provide both image-level anomaly classification (AC) and pixel-level anomaly segmentation (AS) annotations, whereas OCT17, Chest, and HIS are evaluated for image-level anomaly classification only. The labeled pool is used only for source-domain training or few-shot support construction, depending on the evaluation protocol, while the test split is kept fixed for evaluation. Normal-only training splits provided by some datasets are not used under our few-shot or zero-shot protocols. To avoid patient-level leakage, samples from the same patient are assigned to the same split whenever patient/case identifiers are available. For datasets without exposed patient identifiers, we follow the official benchmark splits and do not mix samples across the labeled pool and test set.

\paragraph{BrainMRI.}
The BrainMRI dataset is derived from BraTS2021~\cite{RN21,bakas2017advancing,menze2014multimodal} and uses FLAIR volumes for brain anomaly detection. We extract 2D slices from selected depth ranges of the original 3D volumes and resize them to $240\times240$. The benchmark contains 83 labeled pool samples and 3,715 test images for evaluating both anomaly classification and localization.

\paragraph{LiverCT.}
The LiverCT dataset is constructed from abdominal CT scans, where normal samples are taken from BTCV~\cite{landman2015miccai} and abnormal testing samples are obtained from LiTS~\cite{bilic2023liver}. The CT volumes are converted into grayscale images using an abdominal window and then cropped into 2D axial slices. The benchmark includes 166 labeled pool samples and 1,493 test slices. Since pixel-level lesion annotations are available, LiverCT is used for both AC and AS evaluation.

\paragraph{RESC.}
RESC is a retinal OCT dataset designed for retinal edema analysis. It provides pixel-level annotations of abnormal regions, making it suitable for both image-level and pixel-level anomaly detection. The benchmark includes 115 labeled pool samples and 1,805 test images, with an original image size of $512\times1024$.

\paragraph{OCT17.}
OCT17 is a retinal OCT dataset mainly used for image-level anomaly classification. It contains normal OCT images and multiple abnormal retinal disease categories. The benchmark includes 32 labeled pool samples and 968 test images, with an image size of $512\times496$.

\paragraph{Chest.}
The Chest dataset is based on RSNA chest radiographs~\cite{wang2017chestx,shih2019augmenting}, where abnormal cases are associated with lung opacity or other non-normal findings. The benchmark includes 1,490 labeled pool samples and 17,194 test images. This dataset is evaluated under the image-level anomaly classification setting.

\paragraph{HIS.}
The HIS dataset is constructed from Camelyon16~\cite{ehteshami2017diagnostic}, which consists of hematoxylin and eosin stained histopathology whole-slide images of lymph node tissue. Following the patch-based setting, normal and abnormal image patches are extracted from the original WSIs. The benchmark includes 236 labeled pool samples and 2,000 test patches, and is used for image-level anomaly classification.

\begin{table}[!t]
\centering
\setlength{\tabcolsep}{2.5pt}
\renewcommand{\arraystretch}{1.08}
\resizebox{\linewidth}{!}{
\begin{tabular}{l l c c c c}
\toprule
\textbf{Dataset} & \textbf{Modality} & \textbf{Task} &
\textbf{Pool} & \textbf{Test} & \textbf{Size} \\
\midrule
BrainMRI & Brain MRI & AC+AS & 83 & 3,715 & $240{\times}240$ \\
LiverCT  & Liver CT & AC+AS & 166 & 1,493 & $512{\times}512$ \\
RESC     & Retinal OCT & AC+AS & 115 & 1,805 & $512{\times}1024$ \\
OCT17    & Retinal OCT & AC & 32 & 968 & $512{\times}496$ \\
Chest    & Chest X-ray & AC & 1,490 & 17,194 & $1024{\times}1024$ \\
HIS      & Histopathology & AC & 236 & 2,000 & $256{\times}256$ \\
\bottomrule
\end{tabular}
}
\vspace{1mm}
\caption{\textbf{Summary of the six medical AD datasets.} The labeled pool is used for few-shot support construction, and the test split is used for evaluation.}
\label{tab:dataset_summary}
\end{table}

\section{More Implementation Details}
\label{app:implementation_details}
Here we provide descriptions of more implementation details:
\paragraph{Zero-shot Protocol.}
In the zero-shot setting, no target-domain support images are used for prototype initialization, memory construction, or model adaptation. For each run, one dataset is held out as the unseen target domain, and the remaining datasets are used as source domains. The trained model is then evaluated directly on the held-out test set using the learned visual prototypes, adapters, conditioning modules, and layer-fusion weights.

\paragraph{Few-shot Protocol.}
In the few-shot setting, we construct a small target-domain support set from the labeled pool of the target dataset. For a $K$-shot experiment, the support set contains $K$ normal images and $K$ abnormal images. For datasets with pixel-level annotations, abnormal support images are paired with lesion masks. The support images are used to initialize normal/abnormal visual prototypes, while only normal support images are used to construct the normal-reference memory bank. Unless otherwise specified, we evaluate $K\in\{2,4,8,16\}$ following the same support-shot protocol as prior work.

\begin{table*}[!t]
\centering
\scriptsize
\setlength{\tabcolsep}{2.8pt}
\resizebox{\textwidth}{!}{%
\begin{tabular}{lclccccccc}
\toprule
\multirow{2}{*}{\textbf{Few-shot}} &
\multirow{2}{*}{\textbf{Metric}} &
\multirow{2}{*}{\textbf{Dataset}} &
\textbf{DRA} &
\textbf{APRIL-GAN} &
\textbf{MediCLIP} &
\textbf{MVFA} &
\textbf{MadCLIP} &
\textbf{VisualAD} &
\cellcolor{oursblue}\textbf{ReCAP} \tabularnewline
\cmidrule(lr){4-10}
& & &
N/A &
\xmark &
\xmark &
\xmark &
\xmark &
\cmark &
\cellcolor{oursblue}\cmark \tabularnewline
\midrule

\multirow{9}{*}{$K{=}8$}
& \multirow{6}{*}{\makecell[c]{Image-level\\(AUROC, $F_1$-max, AP)}}
& HIS
& (74.33, 72.10, 73.62)
& (81.70, 76.84, 80.52)
& (69.80, 71.22, 72.42)
& (80.29, 76.34, 78.34)
& (\underline{87.45},  \underline{80.24}, \underline{86.32})
& (64.20, 66.50, 65.10)
& \cellcolor{oursblue}(\textbf{90.01}, \textbf{81.21}, \textbf{90.85}) \tabularnewline

& & Chest
& (82.70, 97.02, 98.34)
& (73.69, 97.22, 98.36)
& (72.08, 96.52, 97.22)
& (75.82, 97.68, 98.23)
& (\underline{83.90}, 97.74, 99.10)
& (62.50, \underline{97.93}, 97.00)
& \cellcolor{oursblue}(\textbf{89.82}, \textbf{98.74}, \textbf{99.02}) \tabularnewline

& & OCT17
& (99.13, 98.02, 99.45)
& (99.75, 98.42, 99.70)
& (95.69, 93.52, 96.82)
& (99.54, 98.41, 99.86)
& (99.14, 98.30, 99.55)
& (\underline{99.80}, 98.60, 99.90)
& \cellcolor{oursblue}(\textbf{99.99}, \textbf{99.72}, \textbf{100.00}) \tabularnewline

& & BrainMRI
& (85.94, 90.14, 94.92)
& (88.41, 92.94, 97.34)
& (92.29, 93.02, 97.32)
& (87.09, 90.94, 97.01)
& (\underline{95.17}, \textbf{95.90}, 99.00)
& (93.90, 93.40, 95.70)
& \cellcolor{oursblue}(\textbf{95.85}, \underline{95.84}, \textbf{98.96}) \tabularnewline

& & LiverCT
& (72.53, 67.24, 73.62)
& (62.38, 65.84, 70.10)
& (86.32, 78.42, 86.24)
& (78.54, 70.67, 80.33)
& (\underline{89.31}, 81.24, 90.22)
& (84.53, 74.00, 84.90)
& \cellcolor{oursblue}(\textbf{90.77}, \textbf{86.41}, \textbf{91.90}) \tabularnewline

& & RESC
& (93.06, 84.20, 91.80)
& (91.36, 83.64, 90.52)
& (88.82, 81.34, 88.24)
& (97.57, \textbf{92.50}, 96.43)
& (\underline{97.16}, 89.20, 96.04)
& (92.20, 82.00, 91.50)
& \cellcolor{oursblue}(\textbf{98.28}, \underline{92.42}, \textbf{97.99}) \tabularnewline

\cmidrule(lr){2-10}

& \multirow{3}{*}{\makecell[c]{Pixel-level\\(AUROC, $F_1$-max, PRO)}}
& BrainMRI
& (75.32, 22.42, 58.40)
& (95.50, 40.20, 81.26)
& (98.02, 43.50, 86.92)
& (95.11, 40.69, 77.89)
& (\underline{98.02}, 46.10, 89.20)
& (86.80, 45.30, 83.86)
& \cellcolor{oursblue}(\textbf{98.12}, \textbf{53.60}, \textbf{90.15}) \tabularnewline

& & LiverCT
& (81.78, 30.42, 60.82)
& (97.56, 49.20, 80.64)
& (98.32, 56.80, 85.52)
& (94.59, 57.30, 73.27)
& (\textbf{99.81}, 60.80, 88.62)
& (99.10, 34.90, 82.68)
& \cellcolor{oursblue}(\underline{99.80}, \textbf{69.28}, \textbf{92.27}) \tabularnewline

& & RESC
& (83.07, 38.24, 58.40)
& (97.36, 68.20, 81.40)
& (95.98, 64.80, 77.20)
& (99.33, \textbf{85.23}, \textbf{96.12})
& (98.85, 72.10, 86.30)
& (98.10, 72.20, 91.06)
& \cellcolor{oursblue}(\textbf{99.58}, \underline{83.38}, \underline{94.85}) \tabularnewline

\midrule

\multirow{9}{*}{$K{=}16$}
& \multirow{6}{*}{\makecell[c]{Image-level\\(AUROC, $F_1$-max, AP)}}
& HIS
& (79.16, 74.52, 78.20)
& (81.16, 78.90, 81.40)
& (70.22, 71.82, 74.00)
& (83.69, 77.55, 82.39)
& (\underline{90.14}, 82.00, 89.50)
& (63.10, 67.20, 60.30)
& \cellcolor{oursblue}(\textbf{90.28}, \textbf{82.17}, \textbf{90.92}) \tabularnewline

& & Chest
& (85.01, 97.10, 98.55)
& (78.62, 97.35, 98.50)
& (69.74, 96.22, 96.82)
& (84.40, 97.69, 98.96)
& (\underline{88.15},  97.80, 99.20)
& (63.80, \textbf{98.12}, 97.00)
& \cellcolor{oursblue}(\textbf{90.19}, \underline{98.07}, \textbf{99.25}) \tabularnewline

& & OCT17
& (99.87, 98.50, 99.85)
& (\underline{99.93}, 98.70, 99.90)
& (96.37, 94.10, 97.20)
& (98.85, 98.96, 99.48)
& (99.71, 98.60, 99.70)
& (99.40, 97.60, 99.80)
& \cellcolor{oursblue}(\textbf{100.00}, \textbf{99.93}, \textbf{100.00}) \tabularnewline

& & BrainMRI
& (82.99, 86.30, 91.50)
& (94.03, 94.20, 98.10)
& (91.56, 92.50, 97.00)
& (94.33, \underline{94.11}, 98.74)
& (95.90, \underline{95.68}, 99.20)
& (\underline{97.10}, 92.90, 96.40)
& \cellcolor{oursblue}(\textbf{97.59}, \textbf{96.21}, \textbf{99.97}) \tabularnewline

& & LiverCT
& (80.89, 72.80, 82.10)
& (82.94, 76.40, 85.30)
& (79.31, 76.20, 84.60)
& (85.35, 75.08, 85.97)
& (\underline{91.46}, 83.20, 91.50)
& (90.70, 82.80, 86.90)
& \cellcolor{oursblue}(\textbf{92.15}, \textbf{88.82}, \textbf{92.13}) \tabularnewline

& & RESC
& (94.88, 86.00, 92.60)
& (95.96, 86.20, 94.00)
& (86.51, 80.00, 86.80)
& (98.81, 94.84, 98.65)
& (\underline{99.11}, 95.50, 99.00)
& (98.80, 85.50, 93.90)
& \cellcolor{oursblue}(\textbf{99.32}, \textbf{95.97}, \textbf{99.13}) \tabularnewline

\cmidrule(lr){2-10}

& \multirow{3}{*}{\makecell[c]{Pixel-level\\(AUROC, $F_1$-max, PRO)}}
& BrainMRI
& (80.45, 25.80, 62.20)
& (96.17, 42.84, 84.20)
& (98.08, 44.10, 87.40)
& (97.45, 40.77, 76.93)
& (97.97, 47.20, 89.80)
& (\textbf{99.50}, 51.70, 85.92)
& \cellcolor{oursblue}(\underline{98.96}, \textbf{58.40}, \textbf{90.36}) \tabularnewline

& & LiverCT
& (93.00, 36.20, 68.40)
& (99.64, 54.80, 86.42)
& (98.95, 58.40, 87.10)
& (99.35, 70.22, 74.51)
& (\underline{99.74}, 64.20, 90.20)
& (99.40, 45.30, 84.56)
& \cellcolor{oursblue}(\textbf{99.75}, \textbf{72.42}, \textbf{95.01}) \tabularnewline

& & RESC
& (84.01, 40.20, 60.80)
& (98.47, 73.40, 85.20)
& (94.07, 66.80, 78.00)
& (99.49, 88.56, \textbf{96.50})
& (\underline{99.10}, 78.00, 91.20)
& (95.00, 76.70, 92.67)
& \cellcolor{oursblue}(\textbf{99.45}, \textbf{89.11}, \underline{95.56}) \tabularnewline

\bottomrule
\end{tabular}%
}
\caption{\textbf{Comparisons with state-of-the-art few-shot anomaly detection methods under different few-shot settings ($K=8,16$).} Image-level results are reported as (AUROC, $F_1$-max, AP), and pixel-level results as (AUROC, $F_1$-max, PRO). \cmark denotes language-free methods, while \xmark \  denotes methods using text prompts, and N/A denotes non-CLIP methods. All values are reported in percentage.}
\label{tab:sota_fewshot_k8_k16_full_metrics}

\end{table*}

\paragraph{Preprocessing and Optimization Settings.}
All input images are converted to RGB and resized to $240\times240$. ReCAP is trained for 50 epochs using Adam with $\beta_1=0.5$, $\beta_2=0.999$, batch size 1. The learning rate is set to $5\times10^{-4}$ for zero-shot training and $1\times10^{-3}$ for few-shot training. For datasets without pixel-level annotations, the segmentation loss is omitted. The default logit scale is $\alpha=100$, and the top-$k$ ratio for image-level aggregation is $10\%$. The memory-fusion weight is set to $\lambda=0.5$ in the few-shot setting. The bounded conditioning strength $\eta$ is set to $0.05$ for both zero-shot and few-shot training, unless otherwise specified.

\paragraph{Support Sampling and Augmentation.}
For each target dataset and shot setting, we use a fixed predefined support set selected from the validation split. All compared methods use identical support indices, and no repeated random sampling is performed, ensuring fair and reproducible comparisons. To reduce overfitting, we apply lightweight geometric augmentation, including small rotations, translations, and horizontal and vertical flips. When pixel-level masks are available, the same transformations are applied to the corresponding masks. Otherwise, zero masks are used only as placeholders, without segmentation supervision.

\paragraph{Computing Resources.}
All experiments are conducted on NVIDIA A100 GPUs. Inference-time measurements are evaluated on the same GPU platform using the same input resolution of $240\times240$.

\section{Additional Quantitative Results}
\label{app:additional_quantitative}
Table~\ref{tab:sota_fewshot_k8_k16_full_metrics} reports additional few-shot results under larger support settings ($K=8,16$). The results show that ReCAP maintains strong performance as the number of support samples increases. Under $K=8$, ReCAP achieves the best image-level AUROC on all six datasets, with particularly strong results on HIS ($90.01\%$), LiverCT ($90.77\%$), and RESC ($98.28\%$). For pixel-level segmentation, ReCAP obtains the best AUROC on BrainMRI and RESC, and remains nearly tied with the best method on LiverCT. Under $K=16$, ReCAP again achieves the best image-level AUROC across all six datasets, reaching $100.00\%$ on OCT17 and $99.32\%$ on RESC. On pixel-level segmentation, ReCAP delivers the best AUROC on LiverCT and competitive results on BrainMRI and RESC, while achieving strong $F_1$-max and PRO scores across all three segmentation datasets. These additional results further confirm that ReCAP benefits consistently from increased target-domain support and remains competitive against both prompt-based and language-free baselines.

\paragraph{Robustness Across Random Seeds.}
Table~\ref{tab:five_seed_auroc} reports the mean and standard deviation over five random seeds under $K\in\{2,4,8,16\}$. ReCAP achieves the best mean AC in 23 of the 24 dataset--shot combinations. For AS, ReCAP obtains the best mean AUROC in 7 of the 12 comparisons and remains competitive in the others. Its standard deviation is at most $0.45$ percentage points across all settings, decreasing to at most $0.30$ and $0.25$ points under $K=8$ and $K=16$, respectively. These results demonstrate that ReCAP maintains consistent performance across random seeds rather than relying on a favorable run.

\section{Additional Visualizations}
\label{app:additional_visualizations}
\begin{figure*}[!t]
    \centering
    \includegraphics[width=\linewidth]{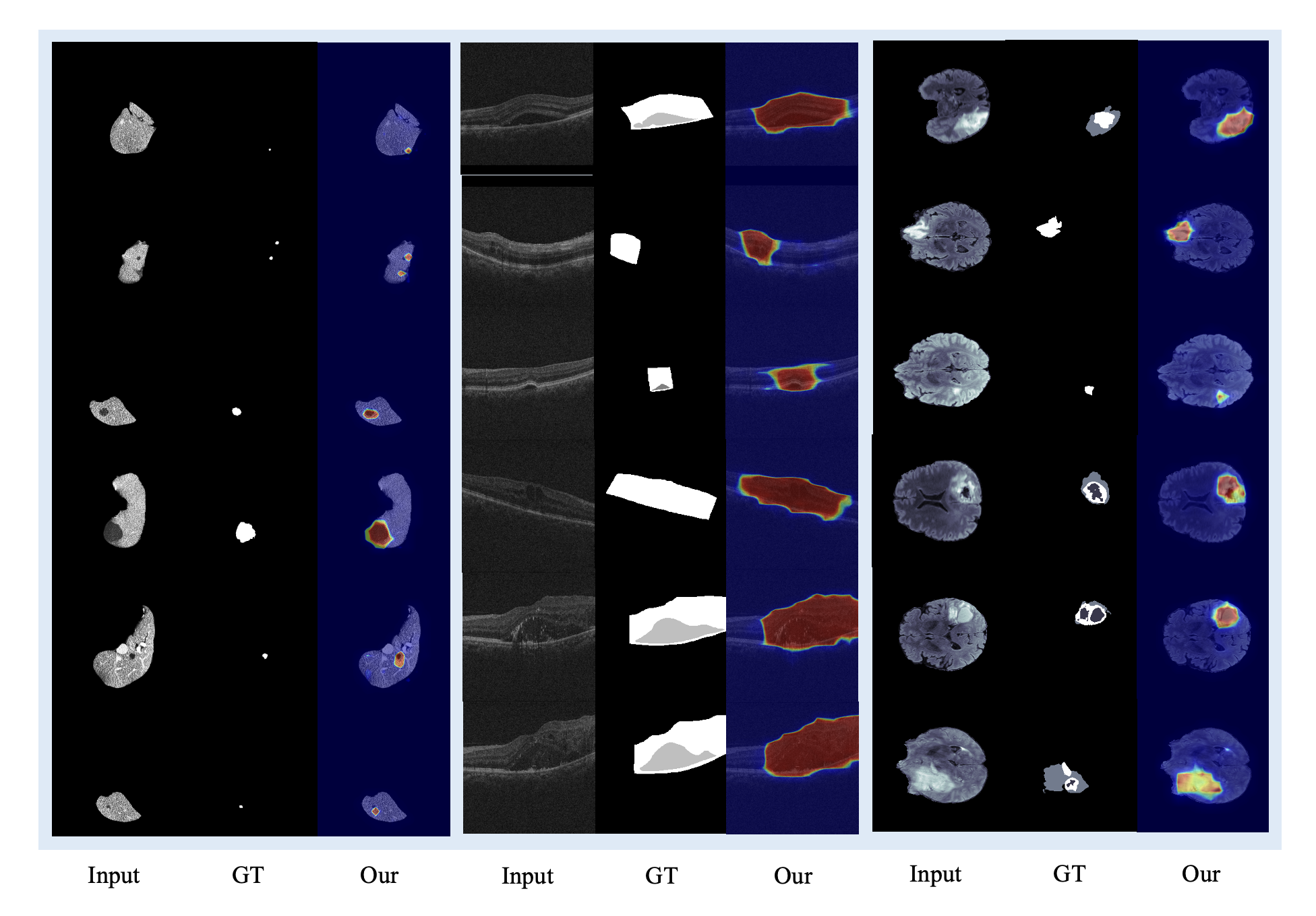}
    \caption{\textbf{Additional qualitative visualizations on different medical anomaly detection datasets.} For each case, we show the input image, ground-truth annotation (GT), and the anomaly map predicted by our method.}
    \label{fig:appendix_visualization}
\end{figure*}

\paragraph{Additional Qualitative Localization.}
Figure~\ref{fig:appendix_visualization} shows additional localization results on LiverCT, RESC, and BrainMRI. ReCAP produces compact anomaly responses that largely overlap with the ground-truth regions while suppressing activations on normal tissues. These examples further show that ReCAP can localize both small focal lesions and larger structural abnormalities across different medical imaging modalities.

\begin{table*}[!t]
\centering
\resizebox{\textwidth}{!}{%
\begin{tabular}{ll l c c c cc cc cc}
\toprule
\multirow{2}{*}{Setting} &
\multirow{2}{*}{Method} &
\multirow{2}{*}{Source} &
\multicolumn{1}{c}{HIS} &
\multicolumn{1}{c}{Chest} &
\multicolumn{1}{c}{OCT17} &
\multicolumn{2}{c}{BrainMRI} &
\multicolumn{2}{c}{LiverCT} &
\multicolumn{2}{c}{RESC} \tabularnewline
\cmidrule(lr){4-4}
\cmidrule(lr){5-5}
\cmidrule(lr){6-6}
\cmidrule(lr){7-8}
\cmidrule(lr){9-10}
\cmidrule(lr){11-12}
& & &
AC & AC & AC &
AC & AS &
AC & AS &
AC & AS \tabularnewline
\midrule

\multirow{7}{*}{$K{=}2$}
& DRA       & CVPR22   & 72.84$\pm$0.86 & 72.36$\pm$0.94 & 98.04$\pm$0.21 & 71.64$\pm$1.12 & 72.31$\pm$1.38 & 57.42$\pm$1.46 & 63.38$\pm$1.72 & 85.57$\pm$0.68 & 65.84$\pm$1.55 \tabularnewline
& APRIL-GAN & CVPRw23  & 69.76$\pm$1.08 & 69.62$\pm$1.02 & \underline{99.18$\pm$0.14} & 78.28$\pm$0.96 & 94.15$\pm$0.64 & 57.64$\pm$1.35 & 96.02$\pm$0.52 & 89.61$\pm$0.74 & 96.24$\pm$0.48 \tabularnewline
& MediCLIP  & MICCAI24 & 64.72$\pm$1.18 & 61.44$\pm$1.30 & 93.52$\pm$0.56 & 85.02$\pm$0.82 & 97.28$\pm$0.36 & 68.62$\pm$1.08 & 97.18$\pm$0.44 & 83.79$\pm$0.86 & 96.16$\pm$0.62 \tabularnewline
& MVFA      & CVPR24   & 79.68$\pm$0.72 & 78.52$\pm$0.84 & 96.61$\pm$0.42 & 91.82$\pm$0.56 & 97.34$\pm$0.32 & \underline{87.12$\pm$0.64} & 96.82$\pm$0.48 & 94.58$\pm$0.42 & 97.02$\pm$0.50 \tabularnewline
& MadCLIP   & MICCAI25 & \underline{83.48$\pm$0.46} & \underline{84.43$\pm$0.39} & 99.03$\pm$0.16 & \underline{93.84$\pm$0.38} & \textbf{97.86$\pm$0.28} & 84.36$\pm$0.58 & \underline{99.34$\pm$0.15} & \underline{95.02$\pm$0.34} & \underline{97.24$\pm$0.42} \tabularnewline
& VisualAD  & CVPR26   & 53.66$\pm$1.42 & 53.92$\pm$1.35 & 92.54$\pm$0.68 & 83.76$\pm$0.76 & 95.12$\pm$0.58 & 84.28$\pm$0.82 & 99.04$\pm$0.24 & 91.36$\pm$0.64 & 97.02$\pm$0.46 \tabularnewline
\rowcolor{oursblue}
\cellcolor{white}
& \textbf{ReCAP} & -- & \textbf{85.76$\pm$0.32} & \textbf{84.83$\pm$0.35} & \textbf{99.36$\pm$0.08} & \textbf{94.28$\pm$0.28} & \underline{97.48$\pm$0.26} & \textbf{90.36$\pm$0.41} & \textbf{99.58$\pm$0.10} & \textbf{95.38$\pm$0.30} & \textbf{98.29$\pm$0.22} \tabularnewline
\midrule

\multirow{7}{*}{$K{=}4$}
& DRA       & CVPR22   & 68.91$\pm$0.92 & 75.62$\pm$0.80 & 99.02$\pm$0.18 & 80.46$\pm$0.82 & 74.95$\pm$1.10 & 59.48$\pm$1.20 & 72.03$\pm$1.32 & 90.74$\pm$0.54 & 77.46$\pm$1.05 \tabularnewline
& APRIL-GAN & CVPRw23  & 76.28$\pm$0.78 & 77.31$\pm$0.72 & \underline{99.39$\pm$0.12} & 89.06$\pm$0.55 & 94.81$\pm$0.52 & 52.86$\pm$1.18 & 96.38$\pm$0.45 & 94.58$\pm$0.38 & 98.06$\pm$0.34 \tabularnewline
& MediCLIP  & MICCAI24 & 70.62$\pm$0.84 & 57.04$\pm$1.26 & 89.21$\pm$0.72 & 83.96$\pm$0.68 & 96.74$\pm$0.42 & 81.68$\pm$0.74 & 98.52$\pm$0.26 & 87.69$\pm$0.76 & 96.78$\pm$0.52 \tabularnewline
& MVFA      & CVPR24   & \underline{80.12$\pm$0.48} & 82.38$\pm$0.56 & 99.34$\pm$0.13 & 88.76$\pm$0.54 & 95.18$\pm$0.44 & 81.94$\pm$0.68 & \underline{99.61$\pm$0.10} & 94.82$\pm$0.42 & 98.66$\pm$0.30 \tabularnewline
& MadCLIP   & MICCAI25 & 79.96$\pm$0.44 & \underline{88.06$\pm$0.31} & 99.32$\pm$0.14 & \textbf{95.18$\pm$0.26} & \textbf{97.84$\pm$0.24} & \underline{82.86$\pm$0.60} & 99.23$\pm$0.16 & \underline{96.54$\pm$0.28} & \underline{98.84$\pm$0.26} \tabularnewline
& VisualAD  & CVPR26   & 67.46$\pm$0.96 & 65.04$\pm$1.10 & 98.42$\pm$0.28 & 77.66$\pm$0.92 & 84.74$\pm$0.80 & 59.58$\pm$1.24 & 98.31$\pm$0.32 & 91.34$\pm$0.58 & 97.38$\pm$0.36 \tabularnewline
\rowcolor{oursblue}
\cellcolor{white}
& \textbf{ReCAP} & -- & \textbf{82.18$\pm$0.36} & \textbf{88.76$\pm$0.28} & \textbf{99.82$\pm$0.05} & \underline{95.06$\pm$0.24} & \underline{97.50$\pm$0.22} & \textbf{87.42$\pm$0.45} & \textbf{99.72$\pm$0.07} & \textbf{96.86$\pm$0.24} & \textbf{99.05$\pm$0.18} \tabularnewline
\midrule

\multirow{7}{*}{$K{=}8$}
& DRA       & CVPR22   & 74.48$\pm$0.70 & 82.54$\pm$0.62 & 99.10$\pm$0.14 & 85.82$\pm$0.58 & 75.46$\pm$0.92 & 72.38$\pm$0.82 & 81.94$\pm$0.86 & 92.94$\pm$0.42 & 83.22$\pm$0.72 \tabularnewline
& APRIL-GAN & CVPRw23  & 81.56$\pm$0.54 & 73.84$\pm$0.78 & 99.71$\pm$0.08 & 88.28$\pm$0.46 & 95.62$\pm$0.38 & 62.52$\pm$0.95 & 97.48$\pm$0.30 & 91.52$\pm$0.52 & 97.48$\pm$0.32 \tabularnewline
& MediCLIP  & MICCAI24 & 69.96$\pm$0.80 & 71.92$\pm$0.86 & 95.82$\pm$0.42 & 92.16$\pm$0.38 & 97.96$\pm$0.20 & 86.18$\pm$0.50 & 98.44$\pm$0.24 & 88.96$\pm$0.62 & 96.12$\pm$0.44 \tabularnewline
& MVFA      & CVPR24   & 80.42$\pm$0.56 & 75.68$\pm$0.66 & 99.49$\pm$0.10 & 87.22$\pm$0.52 & 95.02$\pm$0.36 & 78.68$\pm$0.72 & 94.72$\pm$0.48 & \underline{97.50$\pm$0.22} & \underline{99.26$\pm$0.16} \tabularnewline
& MadCLIP   & MICCAI25 & \underline{87.32$\pm$0.30} & \underline{83.78$\pm$0.36} & 99.20$\pm$0.15 & \underline{95.08$\pm$0.22} & \underline{98.04$\pm$0.19} & \underline{89.22$\pm$0.38} & \textbf{99.78$\pm$0.06} & 97.04$\pm$0.26 & 98.94$\pm$0.24 \tabularnewline
& VisualAD  & CVPR26   & 64.38$\pm$0.92 & 62.34$\pm$0.98 & \underline{99.76$\pm$0.07} & 94.02$\pm$0.34 & 86.62$\pm$0.70 & 84.66$\pm$0.56 & 99.02$\pm$0.18 & 92.38$\pm$0.48 & 98.24$\pm$0.28 \tabularnewline
\rowcolor{oursblue}
\cellcolor{white}
& \textbf{ReCAP} & -- & \textbf{89.86$\pm$0.24} & \textbf{89.64$\pm$0.22} & \textbf{99.96$\pm$0.03} & \textbf{95.76$\pm$0.18} & \textbf{98.10$\pm$0.16} & \textbf{90.62$\pm$0.30} & \underline{99.73$\pm$0.05} & \textbf{98.16$\pm$0.18} & \textbf{99.52$\pm$0.12} \tabularnewline
\midrule

\multirow{7}{*}{$K{=}16$}
& DRA       & CVPR22   & 79.02$\pm$0.56 & 84.86$\pm$0.48 & 99.84$\pm$0.06 & 83.12$\pm$0.62 & 80.62$\pm$0.80 & 80.74$\pm$0.60 & 93.16$\pm$0.52 & 94.72$\pm$0.34 & 84.18$\pm$0.64 \tabularnewline
& APRIL-GAN & CVPRw23  & 81.28$\pm$0.48 & 78.48$\pm$0.62 & \underline{99.91$\pm$0.04} & 93.92$\pm$0.28 & 96.30$\pm$0.30 & 83.08$\pm$0.52 & 99.58$\pm$0.08 & 96.08$\pm$0.28 & 98.56$\pm$0.22 \tabularnewline
& MediCLIP  & MICCAI24 & 70.36$\pm$0.72 & 69.58$\pm$0.78 & 96.52$\pm$0.34 & 91.42$\pm$0.36 & 98.02$\pm$0.18 & 79.48$\pm$0.58 & 99.02$\pm$0.15 & 86.70$\pm$0.56 & 94.22$\pm$0.50 \tabularnewline
& MVFA      & CVPR24   & 83.56$\pm$0.42 & 84.56$\pm$0.44 & 98.92$\pm$0.18 & 94.20$\pm$0.26 & 97.58$\pm$0.22 & 85.48$\pm$0.44 & 99.28$\pm$0.11 & 98.70$\pm$0.18 & \textbf{99.42$\pm$0.09} \tabularnewline
& MadCLIP   & MICCAI25 & \underline{90.02$\pm$0.20} & \underline{88.04$\pm$0.26} & 99.68$\pm$0.08 & 95.82$\pm$0.20 & 97.90$\pm$0.20 & \underline{91.34$\pm$0.32} & \underline{99.70$\pm$0.05} & \underline{99.04$\pm$0.14} & 99.02$\pm$0.16 \tabularnewline
& VisualAD  & CVPR26   & 63.28$\pm$0.86 & 63.64$\pm$0.92 & 99.34$\pm$0.10 & \underline{96.98$\pm$0.18} & \textbf{99.46$\pm$0.07} & 90.56$\pm$0.38 & 99.34$\pm$0.09 & 98.66$\pm$0.20 & 95.18$\pm$0.42 \tabularnewline
\rowcolor{oursblue}
\cellcolor{white}
& \textbf{ReCAP} & -- & \textbf{90.18$\pm$0.18} & \textbf{90.04$\pm$0.20} & \textbf{99.98$\pm$0.03} & \textbf{97.48$\pm$0.14} & \underline{98.88$\pm$0.12} & \textbf{92.02$\pm$0.25} & \textbf{99.72$\pm$0.04} & \textbf{99.24$\pm$0.10} & \underline{99.38$\pm$0.08} \tabularnewline
\bottomrule
\end{tabular}%
}
\caption{\textbf{Comparisons with state-of-the-art few-shot anomaly detection methods under different few-shot settings ($K{=}2,4,8,16$)}. The AUROC (\%, mean$\pm$std over five seeds) for anomaly classification (AC) and anomaly segmentation (AS) are reported. The best mean result is in bold, and the second-best mean result is underlined.}
\label{tab:five_seed_auroc}
\end{table*}

\begin{figure}[ht]
    \centering
    \includegraphics[width=\linewidth]{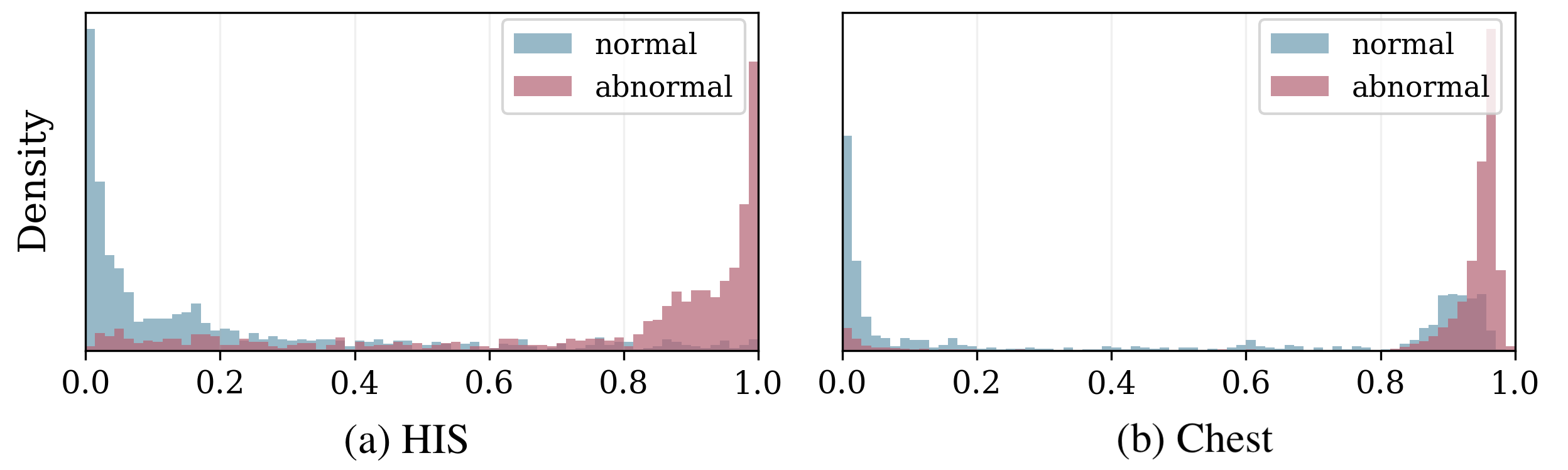}
    \caption{Additional image-level anomaly score distributions under the 16-shot setting. Results are shown for HIS and Chest, the two datasets omitted from the main-paper visualization. }
    \label{fig:score_distribution_full}
\end{figure}

\paragraph{Additional Image-level Anomaly Score Distributions.}
Figure~\ref{fig:score_distribution_full} presents score distributions for HIS and Chest, complementing the four datasets shown in the main paper. On both datasets, normal samples generally receive lower scores, whereas abnormal samples shift toward higher values. HIS exhibits broader distributions and greater overlap, while Chest shows clearer separation with limited overlap at high scores. These results remain consistent with the image-level detection performance. More challenging datasets, such as Histopathology, Chest, and LiverCT, show partial overlap between normal and abnormal distributions. This overlap reflects the difficulty of medical anomaly detection under subtle pathological patterns, imaging noise, and complex anatomical variation. Nevertheless, abnormal samples consistently shift toward higher anomaly scores on all datasets, indicating that ReCAP learns a robust image-level scoring function across heterogeneous medical domains.

\begin{figure}[!ht]
    \centering
    \includegraphics[width=\linewidth]{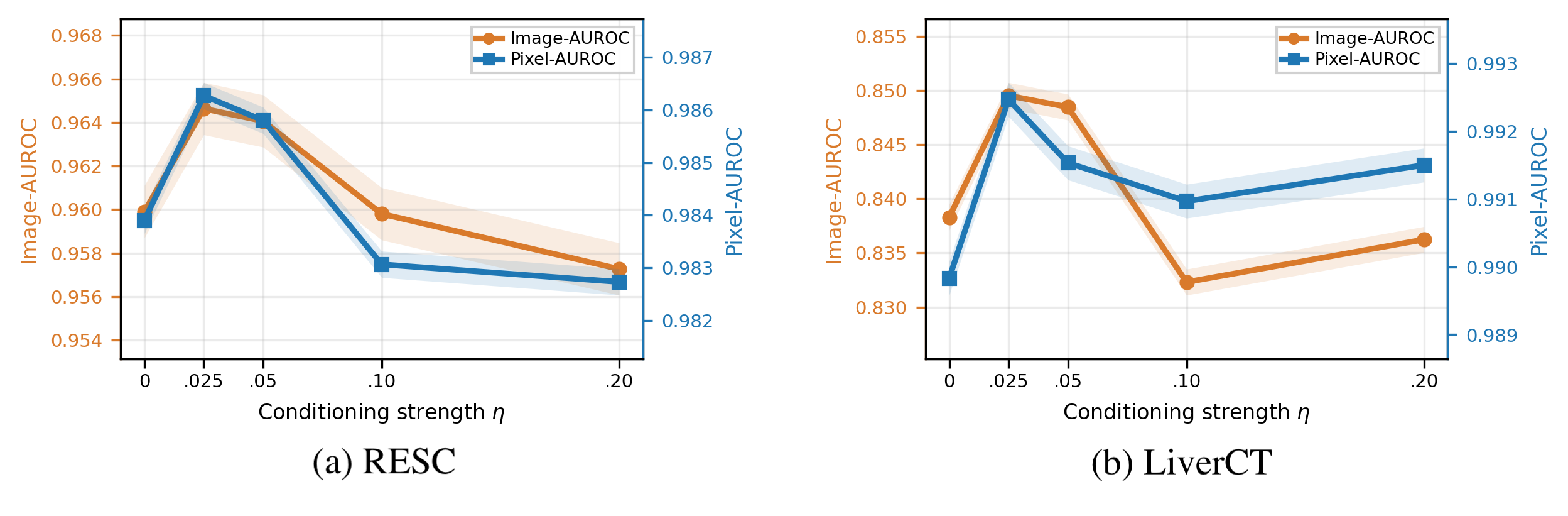}
    \caption{Sensitivity analysis of the bounded conditioning strength $\eta$ on Liver and RESC under the 4-shot setting. Points and error bars denote the mean and standard deviation over three random seeds. Moderate values of $\eta$ provide a better trade-off, while overly large $\eta$ may over-constrain the representation and degrade performance.}
    \label{fig:eta_sensitivity}
\end{figure}

\section{Additional Parameter Analysis}
\label{app:parameter_analysis}

\begin{table}[t]
\centering
\setlength{\tabcolsep}{6pt}
\renewcommand{\arraystretch}{1.12}
\begin{tabular}{llccc}
\toprule
Model & Backbone & Shots & AC & AS \\
\midrule
\multirow{4}{*}{CLIP}
& ViT-B/16       & 2 & 70.56 & 97.59 \\
& ViT-L/14@336px & 2 & \textbf{90.45} & \textbf{99.64} \\
\cmidrule(lr){2-5}
& ViT-B/16       & 4 & 85.32 & 97.98 \\
& ViT-L/14@336px & 4 & 87.56 & \textbf{99.77} \\
\midrule
\multirow{4}{*}{DINOv2}
& ViT-B/14 & 2 & 87.46 & 99.18 \\
& ViT-L/14@336px & 2 & 89.85 & 99.40 \\
\cmidrule(lr){2-5}
& ViT-B/14 & 4 & 88.99 & 99.42 \\
& ViT-L/14@336px& 4 & \textbf{89.20} & 99.54 \\
\bottomrule
\end{tabular}
\caption{Effects of pre-trained visual backbones on LiverCT. All models use $240\times240$ inputs. AC and AS denote image- and pixel-level AUROC (\%), respectively.}
\label{tab:clip_backbone_ablation}
\end{table}

\paragraph{Effects of Pre-trained Visual Backbones.}
Table~\ref{tab:clip_backbone_ablation} examines ReCAP with different pre-trained visual backbones on LiverCT. Among CLIP backbones, ViT-L/14@336px substantially outperforms ViT-B/16 under 2-shot supervision, improving AC from $70.56\%$ to $90.45\%$ and AS from $97.59\%$ to $99.64\%$. More importantly, even when using the lighter ViT-B/16 backbone, ReCAP still outperforms APRIL-GAN, which uses the same backbone scale, on both AC and AS under $K=2$ and $K=4$. This indicates that the gains are not merely due to a larger visual encoder, but also come from the proposed re-centered anomaly prototypes and normal-reference modeling. Under 4-shot supervision, the gap between ViT-B/16 and ViT-L/14@336px becomes smaller, suggesting reduced sensitivity to backbone capacity as target-domain support increases. ReCAP also transfers effectively to DINOv2 backbones: ViT-B/14 achieves $88.99\%$ AC and $99.42\%$ AS with four shots, approaching the performance of ViT-L/14.

\paragraph{Sensitivity to the Bounded Conditioning Strength $\eta$.}
The parameter $\eta$ controls the strength of the bounded conditioning constraint in our method. We therefore study its influence under the 4-shot setting by varying $\eta$ while keeping all other hyperparameters fixed. As shown in Figure~\ref{fig:eta_sensitivity}, moderate values of $\eta$ yield better or comparable mean performance on both Liver and RESC across Image-AUROC, Pixel-AUROC, and PRO. In particular, $\eta=0.025$ and $\eta=0.05$ consistently achieve competitive results, suggesting that a softly bounded conditioning signal can provide useful anatomical guidance without dominating the learned representation. In contrast, larger values such as $\eta=0.1$ and $\eta=0.2$ tend to degrade the performance, especially on localization-related metrics. These results indicate that the bounded conditioning strength should be moderate rather than overly strong.

\begin{figure*}[t]
    \centering
    \includegraphics[width=\textwidth]{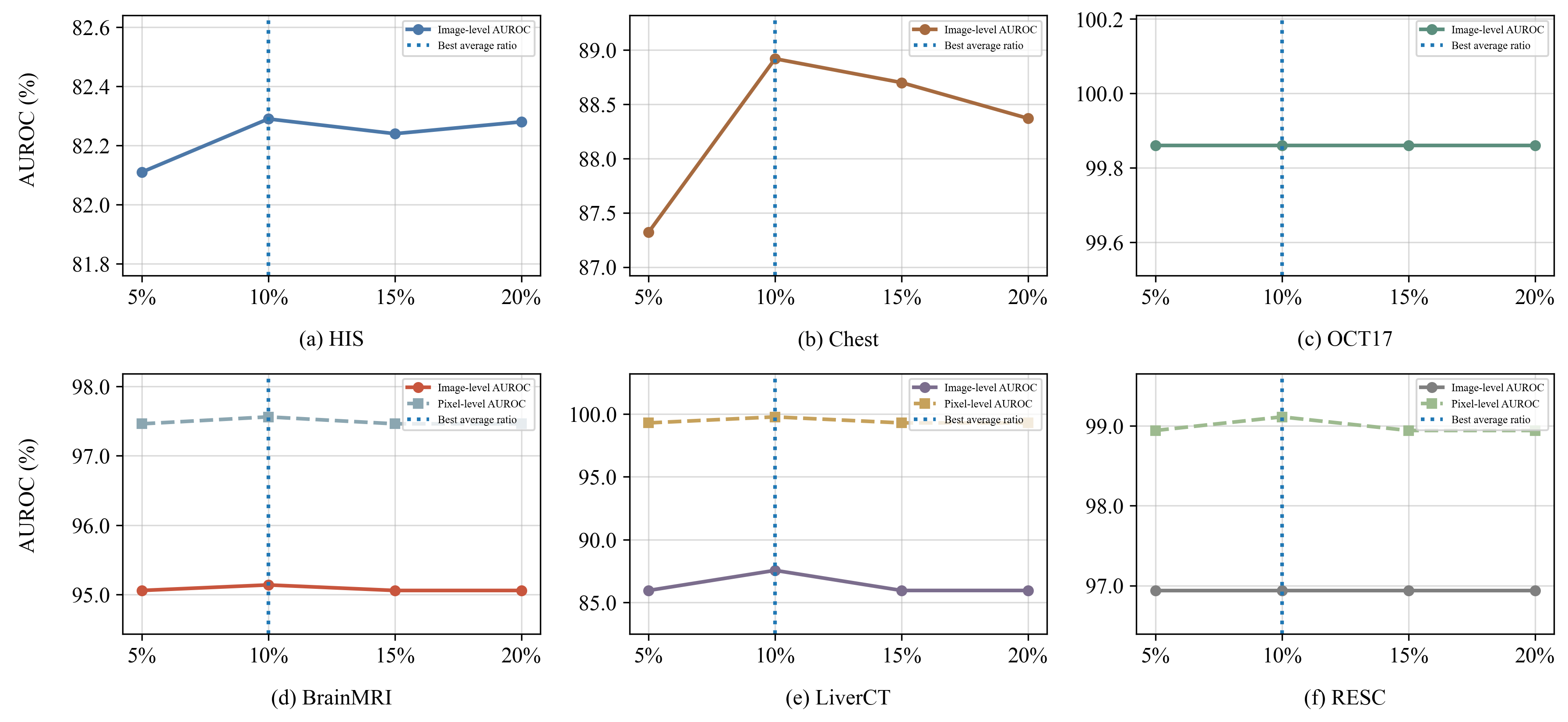}
    \caption{Sensitivity analysis of the top-$k$ ratio under the 4-shot setting. Blue curves report image-level AUROC, and orange dashed curves report pixel-level AUROC when pixel annotations are available. The vertical dotted line marks the default ratio of 10\%, which achieves the best average performance while maintaining stable results across datasets.}
    \label{fig:topk_sensitivity}
\end{figure*}

\begin{figure}[!ht]
    \centering
    \includegraphics[width=0.45\textwidth]{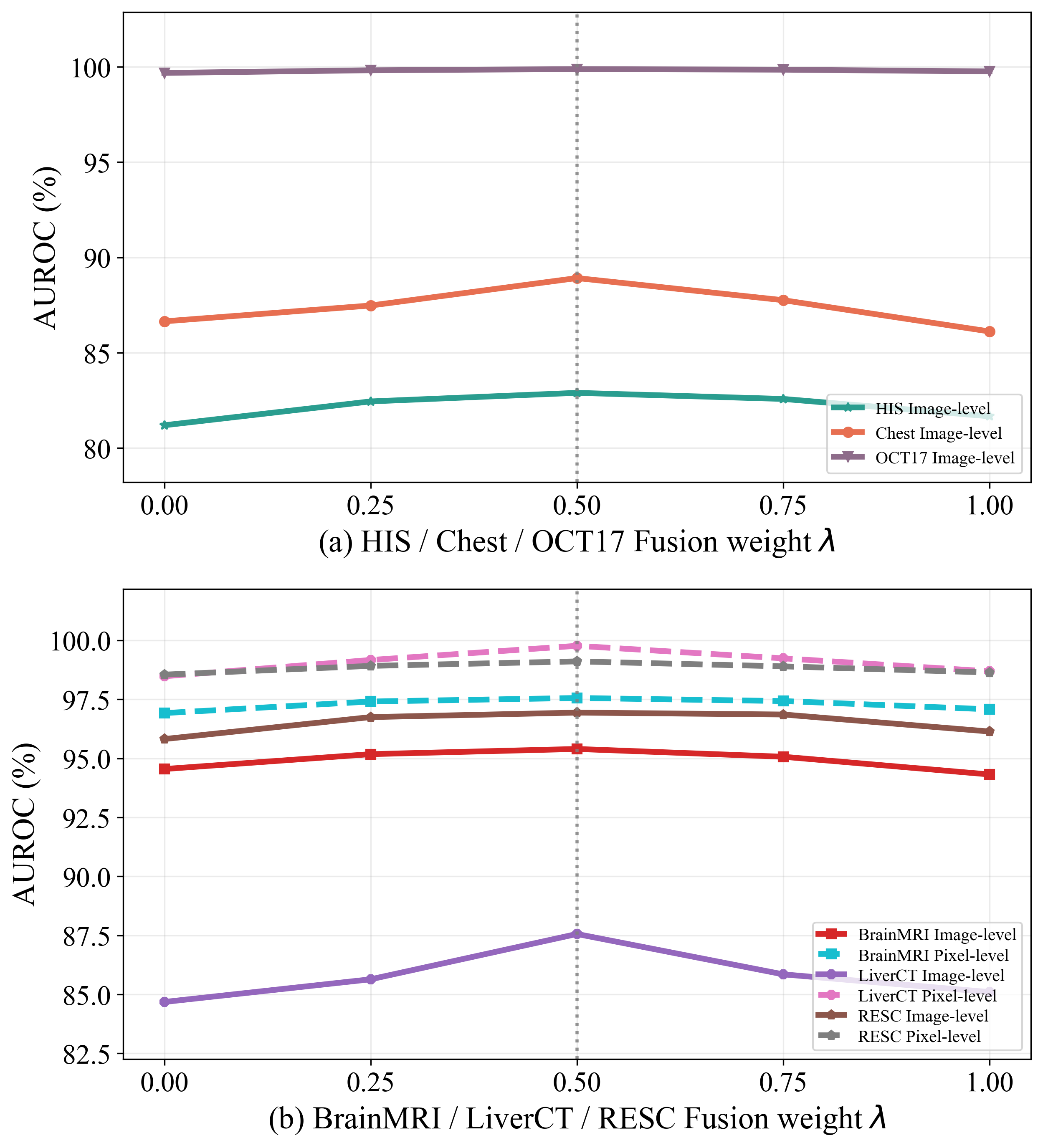}
    \caption{Sensitivity analysis of the fusion weight $\lambda$ in the few-shot setting. For classification-only datasets, only image-level AUROC is reported. For datasets with pixel-level annotations, both image-level AUROC and pixel-level AUROC are shown.}
    \label{fig:lambda_sensitivity}
\end{figure}
\paragraph{Parameter analysis on the top-$k$ ratio.}
We analyze the sensitivity of ReCAP to the top-$k$ ratio used for image-level score aggregation under the 4-shot setting. As shown in Figure~\ref{fig:topk_sensitivity}, performance remains stable when the ratio varies from 5\% to 20\% across all datasets. The best average performance is obtained at 10\%, which is used as the default value in our experiments. A smaller ratio may rely on too few highly activated patches and become sensitive to local noise, whereas a larger ratio may include more normal regions and dilute anomaly evidence. These results show that ReCAP is not sensitive to this hyperparameter within a reasonable range.

\paragraph{Sensitivity to Fusion Weight $\lambda$.}
Figure~\ref{fig:lambda_sensitivity} analyzes the effect of the fusion weight $\lambda$ between the conditional prototype branch and the normal-reference memory branch under the few-shot setting. For classification-only datasets, we report image-level AUROC, while for datasets with pixel-level annotations, we report both image-level and pixel-level AUROC. Across most datasets, the best or near-best performance is achieved around $\lambda=0.5$, indicating that relying solely on either the prototype branch ($\lambda=0$) or the memory branch ($\lambda=1$) is suboptimal. In particular, the segmentation datasets show consistently strong pixel-level AUROC within $\lambda\in[0.25,0.75]$, suggesting that memory-based normal references complement the conditional prototype scores for localization. Overall, the performance curves remain relatively stable around the default value, demonstrating that ReCAP is robust to the exact fusion weight. We therefore set $\lambda=0.5$ in all few-shot experiments.